\documentclass{article}

\usepackage{arxiv}

\usepackage[utf8]{inputenc} % allow utf-8 input
\usepackage[T1]{fontenc}    % use 8-bit T1 fonts
\usepackage{hyperref}       % hyperlinks
\usepackage{url}            % simple URL typesetting
\usepackage{booktabs}       % professional-quality tables
\usepackage{amsfonts}       % blackboard math symbols
\usepackage{amsmath}
\usepackage{nicefrac}       % compact symbols for 1/2, etc.
\usepackage{microtype}      % microtypography
\usepackage{lipsum}		% Can be removed after putting your text content
\usepackage{graphicx}
\usepackage{xcolor}
\usepackage{overpic}
\usepackage{caption}
\usepackage{placeins}
\usepackage{natbib}
\usepackage{doi}

\title{NH-CROP: Robust Pricing for Governed Language Data Assets under Cost Uncertainty}

\author{
	\href{https://orcid.org/0009-0001-9202-7088}{\includegraphics[scale=0.06]{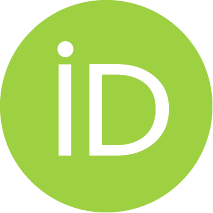}\hspace{1mm}Xu Zheng}
	\hspace{0.75em}
	\href{https://orcid.org/0009-0001-9198-4783}{\includegraphics[scale=0.06]{orcid.pdf}\hspace{1mm}Feiyu Wu}
	\hspace{0.75em}
	\href{https://orcid.org/0009-0009-2355-1191}{\includegraphics[scale=0.06]{orcid.pdf}\hspace{1mm}Zhuocheng Wang}
	\hspace{0.75em}
	\href{https://orcid.org/0009-0001-0732-8223}{\includegraphics[scale=0.06]{orcid.pdf}\hspace{1mm}Yiming Dai}
	\hspace{0.75em}
	\href{https://orcid.org/0000-0001-8310-7169}{\includegraphics[scale=0.06]{orcid.pdf}\hspace{1mm}Hui Li}\thanks{Corresponding author: lihui@mail.xidian.edu.cn} \\
	School of Cyber Engineering, Xidian University \\
	\texttt{zhengxu200477@gmail.com, sn0wm1ans@gmail.com, smilencet1@gmail.com} \\
	\texttt{d18797323123@qq.com, lihui@mail.xidian.edu.cn}
}

\renewcommand{\shorttitle}{\textit{arXiv} Template}

\hypersetup{
pdftitle={NH-CROP: Robust Pricing for Governed Language Data Assets under Cost Uncertainty},
pdfsubject={},
pdfauthor={Xu Zheng, Feiyu Wu, Zhuocheng Wang, Yiming Dai, Hui Li},
pdfkeywords={NH-CROP, Robust Pricing, Governed Language Data Assets, Cost Uncertainty},
}

\newcounter{algorithm}

\begin{document}
\maketitle
\begin{abstract}
Language data are increasingly acquired and governed as assets, yet platforms often price candidate resources before knowing their true privacy or access costs. We study online pricing for governed language data assets under cost uncertainty. At each round, a platform observes an NLP task, a candidate asset, and a coarse cost estimate, may pay for a refined cost signal, posts a price, and receives safe net revenue.

We introduce \textsc{NH-CROP}, a clipped robust pricing framework with a no-harm information-acquisition gate. The method compares direct pricing, risk-aware pricing, and verify-then-price, and acquires information only when its estimated decision value exceeds the best no-verification alternative. Across synthetic, real-proxy, and downstream-utility-grounded benchmarks, clipped \textsc{NH-CROP} variants improve or remain competitive with price-only and risk-aware baselines. Causal ablations show that paid verification is not the main source of gains in real-proxy and utility-grounded settings: the strongest learned policies often choose not to verify. Oracle and high-decision-value diagnostics show that refined cost information can still have substantial local value. Overall, governed language-data platforms should calibrate pricing under uncertain access costs first and verify only when information is cheap and decision-actionable.
\end{abstract}

\section{Introduction}
\label{sec:introduction}

Language data are increasingly treated as governed assets rather than freely interchangeable training inputs. 
Modern NLP systems depend on corpora, instruction data, domain-specific slices, and evaluation resources whose provenance, filtering, documentation, and licensing can affect downstream behavior \citep{bender2018data,gebru2021datasheets,pushkarna2022data,dodge2021documenting,soldaini2024dolma,li2024datacomplm}. 
At the same time, the usefulness of a data asset is task-dependent: a slice that is valuable for domain adaptation, sentiment analysis, or instruction tuning may be less useful for another buyer or model \citep{gururangan2020dont,swayamdipta2020dataset,longpre2023flan,xia2024less}. 
This creates a practical question that is not addressed by dataset documentation or data selection alone: how should a platform price access to a candidate language data asset when both its task value and its privacy/access cost are only partially known?

The cost side is especially difficult. 
A language data asset may carry privacy risk, license restrictions, duplication, benchmark contamination, sensitive content, or quality issues that are not fully visible from coarse metadata \citep{carlini2021extracting,carlini2023quantifying,kandpal2022deduplicating,lee2022deduplicating}. 
A platform can sometimes acquire more information before pricing, for example by inspecting richer documentation, sampling a preview, running a lightweight PII or duplication scan, or performing a small pilot evaluation. 
However, such verification is itself costly. 
The central question is therefore not simply whether the platform is uncertain, but whether reducing that uncertainty would change a consequential pricing decision.

We study this problem as online pricing for governed language data assets under privacy/access-cost uncertainty. 
At round $t$, the platform observes an NLP task context $x_t$, a candidate asset $d_t$, and a coarse cost estimate $\tilde c_t$. 
It may pay a verification cost $c_{\mathrm{ver}}$ to obtain a refined cost signal, posts a price $p_t$, observes binary purchase feedback $y_t$, and receives safe net reward
\begin{equation}
    r_t = y_t(p_t - c_t^\star) - c_{\mathrm{ver}}v_t ,
\end{equation}
where $c_t^\star$ is the true privacy/access cost and $v_t \in \{0,1\}$ is the verification decision. 
Unlike standard dynamic pricing, the objective is not raw revenue but cumulative safe net revenue, which subtracts uncertain access costs and information-acquisition costs. 
This setting connects contextual pricing \citep{besbes2009dynamic,chu2011contextual,ban2021personalized,tullii2024improved}, data pricing \citep{chen2024learning}, and costly information acquisition \citep{bouneffouf2017context,shim2018joint,li2021active,tucker2023bandits}, but differs in that the acquired information concerns the cost side of a governed language-data transaction.

A natural policy is to verify whenever cost uncertainty is high. 
Our empirical audit shows that this is insufficient: verification can reduce cost-estimation error without improving safe net revenue if the refined information does not alter the best pricing action. 
This motivates our key distinction between \emph{cost uncertainty} and \emph{decision value}. 
Cost uncertainty measures how little the platform knows about $c_t^\star$; decision value asks whether knowing more would change the selected price enough to justify paying for information.

\begin{figure*}[t]
    \centering
    \begin{overpic}[width=\textwidth]{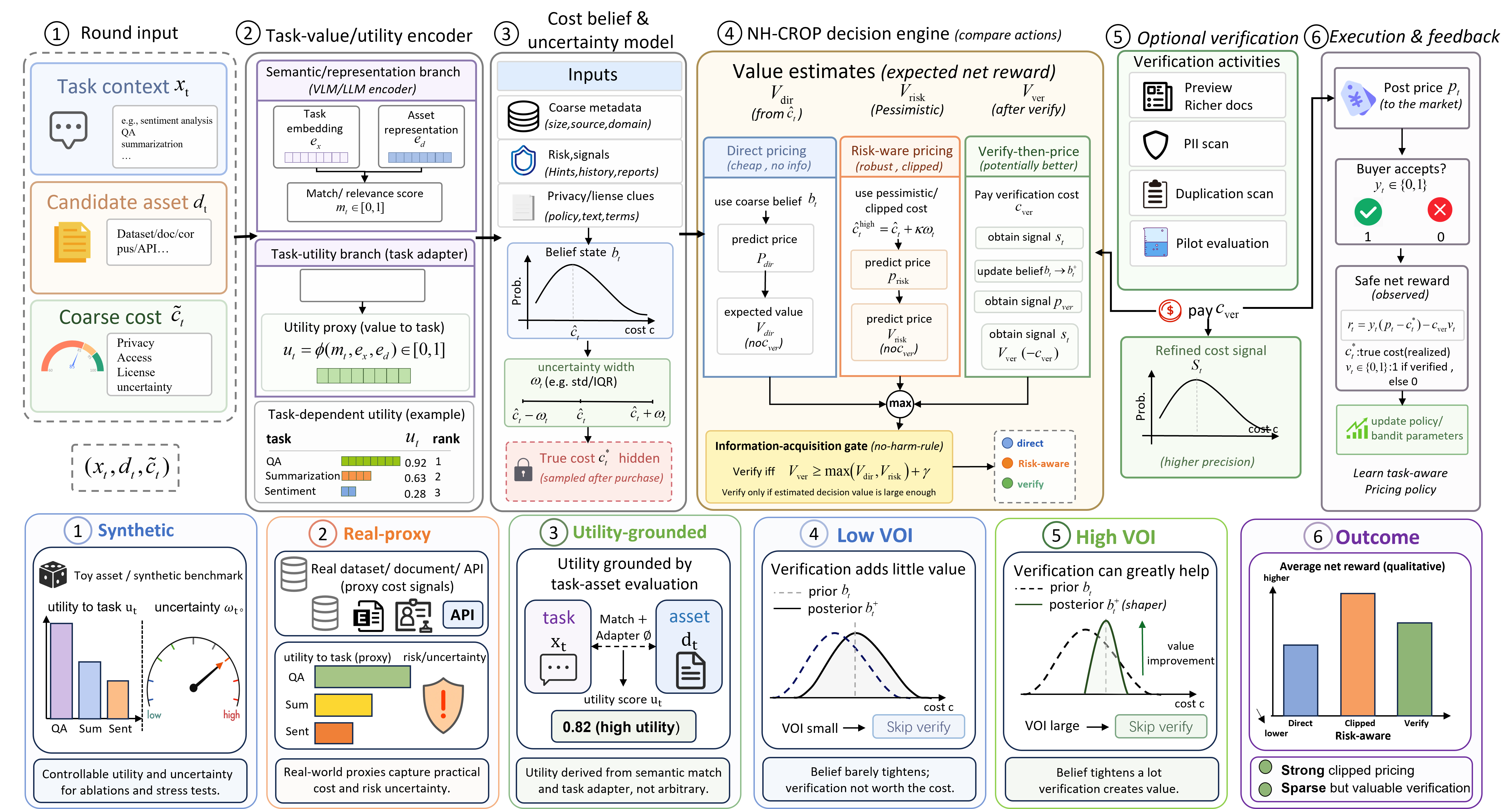}
        \put(45,86){\colorbox{white}{\parbox{0.29\textwidth}{\centering\scriptsize NH-CROP decision engine\\[-0.15em]\scriptsize (compare actions)}}}
    \end{overpic}
    \caption{
    Overview of the NH-CROP pipeline for governed language-data pricing under uncertain privacy/access costs. The figure highlights the task-value and utility encoding components, the cost-belief and uncertainty model, the NH-CROP decision engine that compares direct pricing, risk-aware pricing, and optional verification, and the resulting pricing and information-acquisition outcomes. The bottom-row benchmark, VOI, and outcome panels are illustrative scenarios rather than exact quantitative results.
    }
    \label{fig:overview}
\end{figure*}

We propose \textsc{NH-CROP}, a no-harm clipped robust online pricing framework. 
The method has two components. 
First, it uses clipped optimistic pricing to avoid over-aggressive price choices caused by uncalibrated confidence bonuses under cost uncertainty. 
Second, it treats verification as an optional information-acquisition action: before verifying, the policy compares direct pricing, risk-aware pricing, and verify-then-price, and pays for verification only when the estimated value of refined information exceeds the best no-verification alternative. 
Thus, the method is not designed to verify frequently; it is designed to avoid paying for information when the information is not actionable.
Zero verification can therefore be intended behavior rather than evidence of method failure.
Figure~\ref{fig:overview} provides an overview of the pricing and optional verification workflow studied in this paper.

We evaluate \textsc{NH-CROP} on three benchmark families: a controlled synthetic market, a real-proxy benchmark built from text classification data slices, and a downstream-utility-grounded benchmark where asset value is tied to task performance under lightweight NLP models.
Across these settings, clipped \textsc{NH-CROP} variants improve over Price-Only UCB in all settings and are strongest or competitive among learned non-oracle baselines. 
Causal ablations further show that actual paid verification is not the main source of gains in real-proxy and utility-grounded settings; the strongest learned policies often choose not to verify. 
Oracle analyses nevertheless show that refined cost information can have substantial potential value, indicating that oracle value does not imply learned verification value.

We further stress-test this interpretation with two additional robustness checks.
First, we replace the original lightweight utility matrix with a transformer-derived utility matrix based on \texttt{intfloat/e5-small-v2} (\citealp{wang2022text}).
The resulting utility distribution is weakly correlated with the original lightweight utility distribution, yet the strongest learned policies still avoid paid verification.
Second, we ablate a calibrated VOI gate across high-VOI, real-proxy, and utility-grounded settings.
These checks show that the difficulty is not merely a TF--IDF artifact or a single-threshold overfitting issue: verification becomes useful only in high-decision-value, low-verification-cost regimes, while no-verification robust pricing remains a strong fallback elsewhere.

Our contributions are fourfold. 
First, we formulate online pricing for governed language data assets under uncertain privacy/access costs, with cumulative safe net revenue as the objective. 
Second, we introduce \textsc{NH-CROP}, a clipped robust pricing method with a no-harm information-acquisition gate that compares direct pricing, risk-aware pricing, and verify-then-price before paying for information. 
Third, we provide a causal empirical audit showing that robust pricing calibration, rather than actual paid verification, is the dominant practical driver in our main benchmarks; oracle analyses reveal substantial potential value of refined cost information, but also a gap between oracle value and learnable verification value.
Fourth, we add transformer-utility and CalVOI-ablation robustness checks showing that the no-verification conclusion is not merely an artifact of the original lightweight utility proxy, and that calibrated verification is beneficial only when refined information is both cheap and decision-actionable.
\section{Related Work}
\label{sec:related_work}

\paragraph{Governed language data and data utility.}
NLP datasets are increasingly viewed as governed artifacts whose provenance, documentation, intended use, and limitations affect downstream systems \citep{bender2018data,gebru2021datasheets,mitchell2019model,pushkarna2022data,holland2018dataset,paullada2021data,sambasivan2021everyone}. 
Large-corpus studies similarly show that filtering, mixture design, and documentation are central to language-model development \citep{dodge2021documenting,gao2021pile,soldaini2024dolma,li2024datacomplm}. 
A complementary line of work estimates the utility of data for downstream tasks, including data valuation, influence-style methods, dataset cartography, domain adaptation, and instruction-data selection \citep{ghorbani2019data,ilyas2022datamodels,koh2017understanding,pruthi2020estimating,swayamdipta2020dataset,gururangan2020dont,longpre2023flan,zhou2023lima,xia2024less}. 
These works motivate treating language data as task-dependent assets, but they do not address how a platform should price access online when the asset's privacy or access cost is uncertain.

\paragraph{Dynamic pricing and robust online decisions.}
Dynamic pricing studies how a seller learns demand while repeatedly posting prices \citep{besbes2009dynamic,denboer2015dynamic}, with contextual variants incorporating buyer, product, or market features \citep{li2010contextual,chu2011contextual,abbasi2011improved,agrawal2013thompson,ban2021personalized,javanmard2019dynamic}. 
Recent work also studies pricing data itself and improves algorithms for contextual dynamic pricing \citep{chen2024learning,tullii2024improved}. 
Our setting follows this online-learning perspective, but differs in its cost structure: the platform must optimize safe net revenue while the privacy/access cost of the language asset is only coarsely observed. 
This makes calibration important. 
Work on probability calibration and conservative or safe bandits shows that uncalibrated confidence can harm downstream decisions \citep{niculescu2005predicting,guo2017calibration,kazerouni2017conservative,sui2015safe,amani2019linear}. 
Our clipped pricing rule is related in spirit, but the objective is economic: it limits over-optimistic demand estimates before subtracting uncertain access costs.

\paragraph{Costly information acquisition and verification.}
Our optional verification action is related to active learning, active feature acquisition, and bandits with costly observations \citep{cohn1996active,settles2009active,bouneffouf2017context,shim2018joint,li2021active,tucker2023bandits}. 
The key distinction is the type and timing of the acquired information. 
We do not pay to observe a generic label or reward after acting; instead, the platform may pay before pricing to obtain a refined signal about the cost side of a governed language-data transaction. 
This signal can change the safe margin $p-c$, but it is useful only when it changes a consequential pricing decision. 
This motivates our emphasis on decision value rather than cost-estimation error alone.

\paragraph{Privacy, duplication, and access risk in language data.}
The cost proxies in our benchmarks are motivated by risks that arise in language-data collection and reuse. 
Differential privacy provides a formal disclosure framework \citep{dwork2006calibrating}, while memorization and extraction studies show that training data can sometimes be exposed from language models \citep{carlini2021extracting,carlini2023quantifying}. 
Duplication and near-duplication further affect memorization, contamination, and privacy leakage, and deduplication can improve model behavior and reduce privacy risk \citep{kandpal2022deduplicating,lee2022deduplicating}. 
Our goal is not to propose a new privacy defense, but to study how uncertain privacy/access costs interact with online pricing. 
\textsc{NH-CROP} connects these threads by combining robust pricing calibration with optional no-harm information acquisition for governed language data assets.
\section{Method}
\label{sec:method}

We formulate governed language-data access as an online pricing problem with uncertain privacy/access costs.
At round $t$, the platform observes an NLP task context $x_t$, a candidate language data asset $d_t$, and a coarse cost estimate $\tilde c_t$.
The true cost $c_t^\star$ is hidden before pricing.
The platform may pay verification cost $c_{\mathrm{ver}}$ to obtain a refined cost signal, posts a price $p_t \in \mathcal{P}$, observes purchase feedback $y_t \in \{0,1\}$, and receives safe net reward
\begin{equation}
    r_t = y_t(p_t-c_t^\star)-c_{\mathrm{ver}}v_t ,
    \label{eq:reward}
\end{equation}
where $v_t \in \{0,1\}$ is the verification decision.
The objective is cumulative safe net revenue, not raw revenue.

\paragraph{Cost belief.}
The platform maintains a cost belief $(\mu_t,\sigma_t)$ for the current asset, where $\mu_t$ estimates $c_t^\star$ and $\sigma_t$ captures residual uncertainty.
The coarse estimate and refined verification signal are modeled as
\begin{equation}
    \tilde c_t = c_t^\star+\epsilon_t,
    \qquad
    s_t = c_t^\star+\eta_t ,
\end{equation}
with $\eta_t$ typically lower-variance than $\epsilon_t$.
If verification is skipped, the belief is updated from $\tilde c_t$; if verification is performed, it is updated from $s_t$.
The full belief-update equations, uncertainty floors, and pseudocode are provided in Appendix~\ref{app:formal}.

\paragraph{Contextual demand and safe-revenue score.}
For a candidate price $p$ and cost proxy $c$, we construct a feature vector
\begin{equation}
    \phi_t(p,c)=\phi(x_t,d_t,p,c)
\end{equation}
and estimate purchase probability with a logistic contextual model:
\begin{equation}
    \hat q_t(p,c)
    =
    \sigma_{\mathrm{logit}}
    \left(
    \hat\theta_t^\top \phi_t(p,c)
    \right).
\end{equation}
To encourage exploration, we use a standard contextual-bandit bonus
\begin{equation}
    b_t(p,c)
    =
    \beta_t
    \sqrt{
    \phi_t(p,c)^\top V_t^{-1}\phi_t(p,c)
    } .
\end{equation}
A naive optimistic score can be too aggressive when costs are uncertain, so \textsc{NH-CROP} clips the optimistic purchase estimate:
\begin{equation}
    \bar q_t(p,c)
    =
    \mathrm{clip}
    \left(
    \hat q_t(p,c)+b_t(p,c),0,q_{\max}
    \right),
    \label{eq:clip}
\end{equation}
where $q_{\max}$ is selected on validation seeds and shared by all clipped baselines.
The estimated safe-revenue score is
\begin{equation}
    \widehat R_t(p,c)=\bar q_t(p,c)(p-c).
    \label{eq:safe_score}
\end{equation}

\paragraph{Direct and risk-aware pricing.}
Before considering verification, the platform evaluates two no-verification actions.
The direct action prices from the current cost belief:
\begin{equation}
    V_t^{\mathrm{dir}}
    =
    \max_{p\in\mathcal{P}}
    \widehat R_t(p,\mu_t).
\end{equation}
The risk-aware action prices from a conservative cost proxy:
\begin{equation}
    c_t^{\mathrm{risk}}=\mu_t+\lambda\sigma_t,
    \qquad
    V_t^{\mathrm{risk}}
    =
    \max_{p\in\mathcal{P}}
    \widehat R_t(p,c_t^{\mathrm{risk}}).
\end{equation}
These two alternatives allow the policy to act without paying for information when the current belief is already sufficient.

\paragraph{No-harm information-acquisition gate.}
Verification is considered only if it is expected to improve the final pricing decision.
Let $\mathcal{S}_t$ be the predictive distribution over refined cost signals under the current belief.
Using $K$ Monte Carlo samples $\tilde s_t^{(k)}\sim\mathcal{S}_t$, we estimate the value of verify-then-price as
\begin{equation}
    V_t^{\mathrm{ver}}
    =
    \frac{1}{K}
    \sum_{k=1}^{K}
    \max_{p\in\mathcal{P}}
    \widehat R_t(p,\tilde s_t^{(k)})
    -
    c_{\mathrm{ver}} .
    \label{eq:verify_value}
\end{equation}
\textsc{NH-CROP} verifies only when this estimated value exceeds the best no-verification alternative by margin $\gamma$:
\begin{equation}
    v_t
    =
    \mathbf{1}
    \left[
    V_t^{\mathrm{ver}}
    >
    \max(V_t^{\mathrm{dir}},V_t^{\mathrm{risk}})+\gamma
    \right].
    \label{eq:no_harm_gate}
\end{equation}
If the gate rejects verification, the platform posts the price associated with the better of direct and risk-aware pricing.
If the gate accepts verification, the platform observes $s_t$, updates the cost belief, and prices using the refined estimate.
Thus, zero verification can be the intended behavior when refined information has low estimated decision value.
This conservative gate allows zero verification in regimes where refined cost information is not actionable; in such cases, skipping verification is intended behavior rather than a failure mode.

\paragraph{Relation to uncertainty-triggered verification.}
A simpler baseline verifies whenever uncertainty exceeds a threshold:
\begin{equation}
    v_t^{\mathrm{thr}}=\mathbf{1}[\sigma_t>\tau].
\end{equation}
We refer to this baseline as \textsc{TPIV-UCB}.
TPIV-style rules capture the intuition that high uncertainty should trigger inspection, but they conflate cost uncertainty with decision value.
\textsc{NH-CROP} instead asks whether the refined signal is expected to change the final pricing decision enough to justify its cost.

\paragraph{Decision-value diagnostic.}
For analysis only, we define the counterfactual value of exact cost information:
\begin{equation}
    \Delta_t^{\mathrm{info}}
    =
    \max_{p\in\mathcal{P}}\widehat R_t(p,c_t^\star)
    -
    \max_{p\in\mathcal{P}}\widehat R_t(p,\mu_t).
    \label{eq:info_value}
\end{equation}
This quantity is not observed by the learner because it depends on $c_t^\star$.
It is used to stratify rounds by decision relevance in Section~\ref{sec:experiments}.
A large cost-estimation error does not necessarily imply a large $\Delta_t^{\mathrm{info}}$: verification helps only when refined cost information changes a consequential pricing decision.
\section{Experiments}
\label{sec:experiments}

We evaluate two questions. First, does clipped robust pricing improve safe net revenue under privacy/access-cost uncertainty? Second, when the method performs well, do the gains come from actual paid verification or from robust pricing calibration under coarse cost beliefs? All main results are averaged over 30 random seeds; full benchmark details, hyperparameters, and additional diagnostics are provided in the appendices.

\paragraph{Benchmarks.}
We use three benchmark families. 
\textbf{SYN-high} is a controlled synthetic market with high cost-estimation noise, designed to isolate the interaction between task-conditioned demand, cost uncertainty, and optional information acquisition. 
\textbf{RP-base} and \textbf{RP-high-DV} are real-proxy benchmarks built from SST-2, AG News, and an emotion classification dataset \citep{socher2013recursive,zhang2015character,saravia2018carer}. 
They construct language-data assets from real text slices and derive privacy/access-cost proxies from sensitive-pattern indicators, duplication statistics, quality features, class distributions, and source-level access priors. 
\textbf{UT-base} and \textbf{UT-high} are downstream-utility-grounded benchmarks in which asset value is tied to validation improvement from adding a candidate asset to a small base training set under a TF--IDF logistic-regression model \citep{pedregosa2011scikit}. 
The high-DV/high-tradeoff variants stress settings where cost information is more likely to affect pricing.
Because this utility-grounded setup is intentionally lightweight, Appendix~\ref{app:additional-robustness} adds a transformer-utility sanity check that reconstructs the utility matrix using \texttt{intfloat/e5-small-v2} (\citealp{wang2022text}) representations.
This appendix experiment is not used to tune the main policies; it tests whether the no-verification conclusion persists under a different utility distribution.

\paragraph{Methods.}
We compare Price-Only UCB, Risk-Averse UCB, their clipped variants, \textsc{NH-CROP}, clipped \textsc{NH-CROP}, and a clipped no-verification ablation. 
The latter disables actual verification while keeping the same robust pricing structure, allowing us to test whether gains come from information acquisition or pricing calibration. 
We also report oracle information-acquisition baselines only as upper bounds; they are not deployable. 
The clipping value $q_{\max}=0.8$ is selected on validation seeds and shared by all clipped methods, including Price-Only and Risk-Averse baselines.
Additional robustness diagnostics include calibrated VOI variants, Thompson-sampling-style pricing baselines, and EVSI/estimated-VOI triggers.
We treat these as robustness checks rather than as replacements for the main \textsc{NH-CROP} comparison.

\paragraph{Metrics.}
The primary metric is cumulative safe net revenue,
\[
\sum_{t=1}^{T} [y_t(p_t-c_t^\star)-c_{\mathrm{ver}}v_t].
\]
We also report mean reward per round, verification frequency, realized verification ROI, and seed-level paired comparisons. Main tables report directional paired tests for pre-specified comparisons; appendix diagnostics provide additional robustness summaries where available.

\subsection{Main Result: Robust Clipped Pricing}
\label{subsec:rq1_main_results}

Table~\ref{tab:main_results} reports the main comparison. 
Clipped \textsc{NH-CROP} variants improve over Price-Only UCB in all five original settings and are strongest or competitive among learned non-oracle methods.
We interpret these results as evidence for robust clipped pricing under uncertain access costs, rather than as evidence that paid verification is the primary source of gains.
The gains are significant in SYN-high, RP-base, RP-high-DV, and UT-base; the gain in UT-high is positive but not significant.

The fair clipped baselines do not explain away the result. 
Price-Only and Risk-Averse receive the same clipping opportunity, yet their clipped variants do not systematically match clipped \textsc{NH-CROP}. 
This suggests that clipping is most useful when combined with the no-harm pricing structure, rather than as a generic post-hoc adjustment.

\begin{table*}[t]
\centering
\small
\setlength{\tabcolsep}{4pt}
\begin{tabular}{lrrrrrrrr}
\toprule
Setting
& Price
& Price+Clip
& Risk
& Risk+Clip
& NH
& NH+Clip
& NH+Clip-NoV
& $v$-freq \\
\midrule
SYN-high
& 20.05
& 19.37
& 20.00
& 18.69
& 23.88
& 25.45
& \textbf{25.68}
& 0.026 \\
RP-base
& 35.63
& 34.43
& 36.00
& 34.46
& 37.59
& \textbf{38.01}
& \textbf{38.01}
& 0.000 \\
RP-high-DV
& 20.59
& 19.74
& 20.87
& 19.79
& 22.13
& \textbf{23.42}
& \textbf{23.42}
& 0.000 \\
UT-base
& 4.96
& 4.95
& \textbf{5.45}
& 4.94
& 5.09
& 5.40
& 5.40
& 0.000 \\
UT-high
& 5.08
& 4.61
& 5.02
& 4.77
& 5.13
& \textbf{5.41}
& \textbf{5.41}
& 0.000 \\
\bottomrule
\end{tabular}
\caption{
Main fair-clipped results.
Values are cumulative safe net revenue averaged over 30 seeds.
``Price'' is Price-Only UCB, ``Risk'' is Risk-Averse UCB, ``NH'' is \textsc{NH-CROP}, and ``NH+Clip-NoV'' disables verification while retaining clipped pricing.
The final column reports the verification frequency of clipped \textsc{NH-CROP}.
}
\label{tab:main_results}
\end{table*}

\subsection{Verification and Oracle Information}
\label{subsec:verification_oracle_results}

\begin{figure*}[t]
\centering
\includegraphics[width=\textwidth]{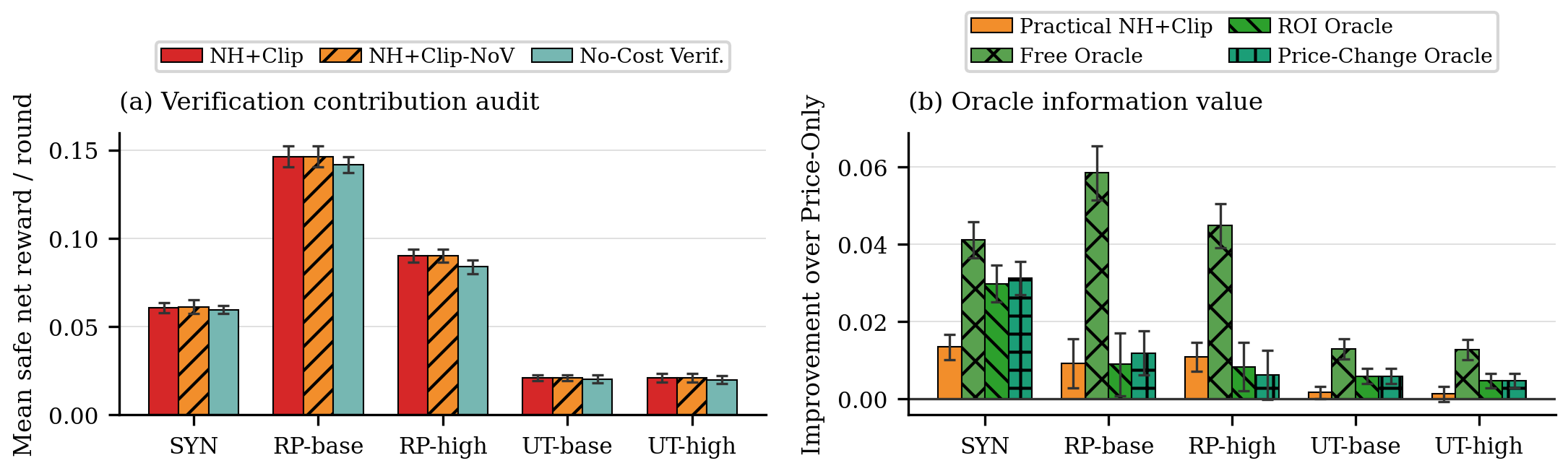}
\caption{
Information-acquisition diagnostics.
(a) The full policy nearly matches its no-verification variant in real-proxy and utility-grounded settings, showing that actual paid verification is not the main source of gains.
(b) Oracle access to refined cost information has substantial value, showing that the difficulty is learning when information is actionable before paying for it. Oracle policies are diagnostic upper bounds and are not deployable.
}
\label{fig:audit_oracle}
\end{figure*}

Figure~\ref{fig:audit_oracle} summarizes two diagnostic audits. 
First, the causal verification audit compares the full policy with a no-verification variant and a no-cost-verification variant. 
If paid verification drove the gains, the full policy should clearly outperform the no-verification version. 
It does not: in real-proxy and utility-grounded settings, the full policy nearly matches the no-verification variant and often verifies zero times. 
Even in SYN-high, it verifies in only 2.6\% of rounds and improves over no-verification by only 0.07 cumulative reward. 
Thus, actual paid verification is not the main empirical driver.

Second, diagnostic oracle upper bounds show that this does not mean cost information is useless. 
A Free Oracle improves over Price-Only UCB by 17.30 in SYN-high, 15.22 in RP-base, and 11.66 in RP-high-DV. 
The gap between oracle information value and learned verification value indicates a harder problem: refined cost information can be valuable, but learned policies do not reliably identify when it is actionable before paying for it.
This is the key empirical distinction of the paper: refined cost information can have oracle value, but a deployable policy must identify useful verification events before paying for them.
In our main benchmarks, the reliable learned behavior is therefore to calibrate pricing first and verify only when the estimated decision value is actionable.

\subsection{Decision Relevance and Verification Events}
\label{subsec:decision_relevance_events}

To analyze when information matters, we stratify rounds by a method-independent decision-relevance score computed from the Price-Only UCB trajectory: the counterfactual value of replacing the current cost belief $\mu_t$ with the true cost $c_t^\star$. 
This analysis partially supports the decision-relevance hypothesis. 
In real-proxy and synthetic settings, clipped \textsc{NH-CROP} variants are strongest in many medium- and high-relevance buckets, while utility-grounded buckets are weaker and less consistent. 
We therefore treat decision relevance as an explanatory diagnostic rather than a complete predictive theory; full bucketed results are in Appendix~\ref{app:decision-relevance}.

We also inspect individual verification events. 
No-cost-verification runs contain many rounds where refined information changes the selected price and yields positive realized ROI, especially in synthetic and real-proxy settings. 
However, learned estimated-VOI policies over-verify and trigger many negative-ROI inspections. 
This reinforces the central conclusion: the bottleneck is not the absence of useful information, but the difficulty of identifying useful information before paying for it. 
Representative cases and full ROI summaries are in Appendix~\ref{app:verification-audit}.

\subsection{Additional Robustness Checks}
\label{subsec:additional_robustness_checks}

Appendix~\ref{app:additional-robustness} reports two additional robustness checks. First, we reconstruct the utility matrix using \texttt{intfloat/e5-small-v2} (\citealp{wang2022text}) representations. The resulting transformer-derived utility distribution is weakly correlated with the original lightweight utility matrix, but the pricing conclusion is unchanged: NH+Clip and NH+Clip-NoV coincide with zero verification, and calibrated verification does not outperform the no-verification fallback. Second, we ablate a calibrated value-of-information gate. Here, CalVOI denotes an appendix-only calibrated value-of-information diagnostic: a validation-trained gate that uses pre-verification features to predict whether verification is likely to have positive realized ROI. It is not part of the primary \textsc{NH-CROP} comparison. CalVOI has a positive window in a high-VOI, low-verification-cost setting, but is unstable at higher verification cost and does not improve over NoV in RP-base or the supplementary embedding-utility settings.
These results reinforce the main claim that verification should be treated as conditional and decision-value-dependent, not as a default response to uncertainty.
\section{Discussion}
\label{sec:discussion}

Our results suggest a more cautious view of information acquisition for governed language data assets. 
Cost uncertainty alone does not justify verification: a refined signal is useful only when it changes a pricing decision enough to improve safe net revenue. 
This distinction explains why uncertainty-triggered verification can fail. 

A platform may reduce cost-estimation error without changing the selected price, or it may change the price in a way that does not improve realized reward. 
Accordingly, \textsc{NH-CROP} treats verification as optional and decision-value-dependent rather than as the default response to uncertainty.

The most reliable component of \textsc{NH-CROP} is clipped robust pricing. 
In contextual pricing, optimism encourages exploration, but uncalibrated optimism can be harmful when the margin $p-c$ depends on an uncertain privacy/access cost. 
Clipping the optimistic purchase estimate limits overly aggressive price choices. 
The fair-clipping comparison in Table~\ref{tab:main_results} shows that this is not merely a post-hoc advantage given only to our method: Price-Only and Risk-Averse baselines receive the same clipping opportunity, yet clipped \textsc{NH-CROP} variants remain strongest or competitive across settings. 
This suggests that robust calibration and the no-harm fallback structure work together.

The additional robustness checks sharpen this interpretation.
The transformer-utility sanity check replaces the original lightweight utility construction with a utility matrix derived from \texttt{intfloat/e5-small-v2} (\citealp{wang2022text}) representations.
The resulting utility values differ substantially from the original lightweight utility distribution, yet the strongest learned policies still avoid paid verification.
This suggests that the no-verification conclusion is not simply a TF--IDF artifact.
However, it also means that transformer-derived utility does not rescue paid verification: robust pricing remains the practical driver.

The verification audit in Figure~\ref{fig:audit_oracle} clarifies the role of information acquisition. 
Actual paid verification is not the main empirical driver in the real-proxy and downstream-utility-grounded settings; the full policy often matches its no-verification variant and frequently chooses not to verify. 
This does not mean that cost information is useless. 
Oracle information-acquisition baselines show substantial potential value, indicating a gap between the \emph{oracle value} of refined cost information and the \emph{learnable value} of practical verification policies. 
Closing this gap likely requires better calibrated value-of-information estimators.

The CalVOI ablations further show that learning actionable verification remains difficult.
In the high-VOI, low-verification-cost setting, calibrated gates can outperform the no-verification fallback, indicating that useful verification regimes exist.
But the advantage disappears or reverses in RP-base, the supplementary embedding-utility settings, and higher-verification-cost settings.
Cross-setting threshold transfer does not systematically fix the issue.
Thus, the bottleneck is not only threshold overfitting; it is the instability of pre-verification signals in regimes where information is weakly actionable.

This has practical implications for language-data platforms. 
Governance risks such as privacy, duplication, licensing, and contamination should be documented and audited \citep{carlini2021extracting,kandpal2022deduplicating,lee2022deduplicating}, but not every reduction in uncertainty affects a pricing or acquisition decision. 
A platform should therefore first calibrate pricing under coarse cost beliefs, then acquire additional information only when it is likely to be actionable.

\section{Conclusion}
\label{sec:conclusion}

We studied robust online pricing for governed language data assets under uncertain privacy/access costs. 
The central lesson is that cost uncertainty is not the same as decision value: verification is useful only when refined information changes a consequential pricing decision enough to justify its cost. 
\textsc{NH-CROP} combines clipped robust pricing with a no-harm information-acquisition gate. 
Across synthetic, real-proxy, and downstream-utility-grounded settings, the most reliable practical gains come from clipped pricing calibration, while actual paid verification contributes only in limited high-decision-value regimes. 
Oracle analyses show that refined cost information can have substantial potential value, but our verification and CalVOI audits show that learning when this information is actionable remains difficult. 
Additional transformer-utility checks support the same qualitative conclusion under a different utility distribution. 
Overall, governed language-data platforms should calibrate pricing under uncertain costs first and acquire additional information only when its decision value is actionable.

\section{Limitations}
\label{sec:limitations}

Our study has several limitations. First, the real-proxy benchmark uses genuine language-data slices, but privacy/access costs are still proxy variables derived from text and metadata rather than deployed marketplace, legal, or contractual costs. Second, although Appendix~\ref{app:transformer-utility} adds a transformer-derived utility sanity check, our utility-grounded experiments remain small-scale and should not be interpreted as large-scale fine-tuning, retrieval-augmented generation, or instruction-tuning evaluations. Third, buyer behavior is simplified to binary purchase feedback rather than strategic negotiation, repeated bargaining, budgeted bundle purchase, or buyer-specific utility. Fourth, verification is modeled primarily as a binary action, whereas real governance workflows may involve staged inspection, legal review, and heterogeneous audit costs. Finally, we do not provide a regret bound for the full clipped no-harm policy. A theoretical analysis of robust pricing with uncertain costs and optional information acquisition remains future work.

\bibliographystyle{unsrtnat}
\bibliography{references}

@article{bender2018data,
  title = {Data Statements for Natural Language Processing: Toward Mitigating System Bias and Enabling Better Science},
  author = {Bender, Emily M. and Friedman, Batya},
  journal = {Transactions of the Association for Computational Linguistics},
  volume = {6},
  pages = {587--604},
  year = {2018}
}

@article{gebru2021datasheets,
  title = {Datasheets for Datasets},
  author = {Gebru, Timnit and Morgenstern, Jamie and Vecchione, Briana and Wortman Vaughan, Jennifer and Wallach, Hanna and Daum{\'e} III, Hal and Crawford, Kate},
  journal = {Communications of the ACM},
  volume = {64},
  number = {12},
  pages = {86--92},
  year = {2021}
}

@inproceedings{mitchell2019model,
  title = {Model Cards for Model Reporting},
  author = {Mitchell, Margaret and Wu, Simone and Zaldivar, Andrew and Barnes, Parker and Vasserman, Lucy and Hutchinson, Ben and Spitzer, Elena and Raji, Inioluwa Deborah and Gebru, Timnit},
  booktitle = {Proceedings of the Conference on Fairness, Accountability, and Transparency},
  pages = {220--229},
  year = {2019}
}

@inproceedings{pushkarna2022data,
  title = {Data Cards: Purposeful and Transparent Dataset Documentation for Responsible AI},
  author = {Pushkarna, Mahima and Zaldivar, Andrew and Kjartansson, Oddur},
  booktitle = {Proceedings of the 2022 ACM Conference on Fairness, Accountability, and Transparency},
  pages = {1776--1826},
  year = {2022}
}

@article{holland2018dataset,
  title = {The Dataset Nutrition Label: A Framework to Drive Higher Data Quality Standards},
  author = {Holland, Sarah and Hosny, Ahmed and Newman, Sarah and Joseph, Joshua and Chmielinski, Kasia},
  journal = {arXiv preprint arXiv:1805.03677},
  year = {2018}
}

@article{paullada2021data,
  title = {Data and Its (Dis)contents: A Survey of Dataset Development and Use in Machine Learning Research},
  author = {Paullada, Amandalynne and Raji, Inioluwa Deborah and Bender, Emily M. and Denton, Emily and Hanna, Alex},
  journal = {Patterns},
  volume = {2},
  number = {11},
  pages = {100336},
  year = {2021}
}

@inproceedings{sambasivan2021everyone,
  title = {``Everyone Wants to Do the Model Work, Not the Data Work'': Data Cascades in High-Stakes AI},
  author = {Sambasivan, Nithya and Kapania, Shivani and Higham, Hannah and Akrong, Diana and Paritosh, Praveen and Aroyo, Lora M.},
  booktitle = {Proceedings of the 2021 CHI Conference on Human Factors in Computing Systems},
  pages = {1--15},
  year = {2021}
}

@inproceedings{dodge2021documenting,
  title = {Documenting Large Webtext Corpora: A Case Study on the Colossal Clean Crawled Corpus},
  author = {Dodge, Jesse and Sap, Maarten and Marasovic, Ana and Agnew, William and Ilharco, Gabriel and Groeneveld, Dirk and Mitchell, Margaret and Gardner, Matt},
  booktitle = {Proceedings of the 2021 Conference on Empirical Methods in Natural Language Processing},
  pages = {1286--1305},
  year = {2021}
}

@article{gao2021pile,
  title = {The Pile: An 800GB Dataset of Diverse Text for Language Modeling},
  author = {Gao, Leo and Biderman, Stella and Black, Sid and Golding, Laurence and Hoppe, Travis and Foster, Charles and Phang, Jason and He, Horace and Thite, Anish and Nabeshima, Noa and others},
  journal = {arXiv preprint arXiv:2101.00027},
  year = {2021}
}

@inproceedings{soldaini2024dolma,
  title = {Dolma: An Open Corpus of Three Trillion Tokens for Language Model Pretraining Research},
  author = {Soldaini, Luca and Kinney, Rodney and Bhagia, Akshita and Schwenk, Dustin and Atkinson, David and Authur, Russell and Bogin, Ben and Chandu, Khyathi and Dumas, Jennifer and Elazar, Yanai and others},
  booktitle = {Proceedings of the 62nd Annual Meeting of the Association for Computational Linguistics},
  pages = {15725--15788},
  year = {2024}
}

@article{li2024datacomplm,
  title = {DataComp-LM: In Search of the Next Generation of Training Sets for Language Models},
  author = {Li, Jeffrey and Fang, Alex and Smyrnis, Georgios and Ivgi, Maor and Jordan, Matt and Gadre, Samir and Bansal, Hritik and Guha, Etash and Keh, Sedrick and Arora, Kushal and others},
  journal = {arXiv preprint arXiv:2406.11794},
  year = {2024}
}

@inproceedings{gururangan2020dont,
  title = {Don't Stop Pretraining: Adapt Language Models to Domains and Tasks},
  author = {Gururangan, Suchin and Marasovic, Ana and Swayamdipta, Swabha and Lo, Kyle and Beltagy, Iz and Downey, Doug and Smith, Noah A.},
  booktitle = {Proceedings of the 58th Annual Meeting of the Association for Computational Linguistics},
  pages = {8342--8360},
  year = {2020}
}

@inproceedings{swayamdipta2020dataset,
  title = {Dataset Cartography: Mapping and Diagnosing Datasets with Training Dynamics},
  author = {Swayamdipta, Swabha and Schwartz, Roy and Lourie, Nicholas and Wang, Yizhong and Hajishirzi, Hannaneh and Smith, Noah A. and Choi, Yejin},
  booktitle = {Proceedings of the 2020 Conference on Empirical Methods in Natural Language Processing},
  pages = {9275--9293},
  year = {2020}
}

@inproceedings{longpre2023flan,
  title = {The FLAN Collection: Designing Data and Methods for Effective Instruction Tuning},
  author = {Longpre, Shayne and Hou, Le and Vu, Tu and Webson, Albert and Chung, Hyung Won and Tay, Yi and Zhou, Denny and Le, Quoc V. and Zoph, Barret and Wei, Jason and others},
  booktitle = {Proceedings of the 40th International Conference on Machine Learning},
  pages = {22631--22648},
  year = {2023}
}

@inproceedings{zhou2023lima,
  title = {LIMA: Less Is More for Alignment},
  author = {Zhou, Chunting and Liu, Pengfei and Xu, Puxin and Iyer, Srinivasan and Sun, Jiao and Mao, Yuning and Ma, Xuezhe and Efrat, Avia and Yu, Ping and Yu, Lili and others},
  booktitle = {Advances in Neural Information Processing Systems},
  year = {2023}
}

@inproceedings{xia2024less,
  title = {LESS: Selecting Influential Data for Targeted Instruction Tuning},
  author = {Xia, Mengzhou and Malladi, Sadhika and Gururangan, Suchin and Arora, Sanjeev and Chen, Danqi},
  booktitle = {Proceedings of the 41st International Conference on Machine Learning},
  pages = {54104--54132},
  year = {2024}
}

@inproceedings{ghorbani2019data,
  title = {Data Shapley: Equitable Valuation of Data for Machine Learning},
  author = {Ghorbani, Amirata and Zou, James},
  booktitle = {Proceedings of the 36th International Conference on Machine Learning},
  pages = {2242--2251},
  year = {2019}
}

@inproceedings{ilyas2022datamodels,
  title = {Datamodels: Predicting Predictions from Training Data},
  author = {Ilyas, Andrew and Park, Sung Min and Engstrom, Logan and Leclerc, Guillaume and Madry, Aleksander},
  booktitle = {Proceedings of the 39th International Conference on Machine Learning},
  year = {2022}
}

@inproceedings{carlini2021extracting,
  title = {Extracting Training Data from Large Language Models},
  author = {Carlini, Nicholas and Tram{\`e}r, Florian and Wallace, Eric and Jagielski, Matthew and Herbert-Voss, Ariel and Lee, Katherine and Roberts, Adam and Brown, Tom B. and Song, Dawn and Erlingsson, {\'U}lfar and others},
  booktitle = {30th USENIX Security Symposium},
  pages = {2633--2650},
  year = {2021}
}

@inproceedings{carlini2023quantifying,
  title = {Quantifying Memorization Across Neural Language Models},
  author = {Carlini, Nicholas and Ippolito, Daphne and Jagielski, Matthew and Lee, Katherine and Tram{\`e}r, Florian and Zhang, Chiyuan},
  booktitle = {International Conference on Learning Representations},
  year = {2023}
}

@inproceedings{kandpal2022deduplicating,
  title = {Deduplicating Training Data Mitigates Privacy Risks in Language Models},
  author = {Kandpal, Nikhil and Wallace, Eric and Raffel, Colin},
  booktitle = {Proceedings of the 39th International Conference on Machine Learning},
  pages = {10697--10707},
  year = {2022}
}

@inproceedings{lee2022deduplicating,
  title = {Deduplicating Training Data Makes Language Models Better},
  author = {Lee, Katherine and Ippolito, Daphne and Nystrom, Andrew and Zhang, Chiyuan and Eck, Douglas and Callison-Burch, Chris and Carlini, Nicholas},
  booktitle = {Proceedings of the 60th Annual Meeting of the Association for Computational Linguistics},
  pages = {8424--8445},
  year = {2022}
}

@article{besbes2009dynamic,
  title = {Dynamic Pricing without Knowing the Demand Function: Risk Bounds and Near-Optimal Algorithms},
  author = {Besbes, Omar and Zeevi, Assaf},
  journal = {Operations Research},
  volume = {57},
  number = {6},
  pages = {1407--1420},
  year = {2009}
}

@article{denboer2015dynamic,
  title = {Dynamic Pricing and Learning: Historical Origins, Current Research, and New Directions},
  author = {den Boer, Arnoud V.},
  journal = {Surveys in Operations Research and Management Science},
  volume = {20},
  number = {1},
  pages = {1--18},
  year = {2015}
}

@inproceedings{chu2011contextual,
  title = {Contextual Bandits with Linear Payoff Functions},
  author = {Chu, Wei and Li, Lihong and Reyzin, Lev and Schapire, Robert E.},
  booktitle = {Proceedings of the Fourteenth International Conference on Artificial Intelligence and Statistics},
  pages = {208--214},
  year = {2011}
}

@article{ban2021personalized,
  title = {Personalized Dynamic Pricing with Machine Learning: High-Dimensional Features and Heterogeneous Elasticity},
  author = {Ban, Gah-Yi and Keskin, N. Bora},
  journal = {Management Science},
  volume = {67},
  number = {9},
  pages = {5549--5568},
  year = {2021}
}

@inproceedings{chen2024learning,
  title = {Learning to Price Homogeneous Data},
  author = {Chen, Keran and Huh, Joon Suk and Kandasamy, Kirthevasan},
  booktitle = {Advances in Neural Information Processing Systems},
  year = {2024}
}

@inproceedings{tullii2024improved,
  title = {Improved Algorithms for Contextual Dynamic Pricing},
  author = {Tullii, Matilde and Gaucher, Solenne and Merlis, Nadav and Perchet, Vianney},
  booktitle = {Advances in Neural Information Processing Systems},
  year = {2024}
}

@inproceedings{tucker2023bandits,
  title = {Bandits with Costly Reward Observations},
  author = {Tucker, Aaron D. and Biddulph, Caleb and Wang, Claire and Joachims, Thorsten},
  booktitle = {Proceedings of the Thirty-Ninth Conference on Uncertainty in Artificial Intelligence},
  pages = {2147--2156},
  year = {2023}
}

@inproceedings{bouneffouf2017context,
  title = {Context Attentive Bandits: Contextual Bandit with Restricted Context},
  author = {Bouneffouf, Djallel and Rish, Irina and Cecchi, Guillermo and F{\'e}raud, Rapha{\"e}l},
  booktitle = {Proceedings of the Twenty-Sixth International Joint Conference on Artificial Intelligence},
  pages = {1468--1475},
  year = {2017}
}

@inproceedings{shim2018joint,
  title = {Joint Active Feature Acquisition and Classification with Variable-Size Set Encoding},
  author = {Shim, Hajin and Hwang, Sung Ju and Yang, Eunho},
  booktitle = {Advances in Neural Information Processing Systems},
  pages = {1375--1385},
  year = {2018}
}

@article{wang2022text,
  title={Text embeddings by weakly-supervised contrastive pre-training},
  author={Wang, Liang and Yang, Nan and Huang, Xiaolong and Jiao, Binxing and Yang, Linjun and Jiang, Daxin and Majumder, Rangan and Wei, Furu},
  journal={arXiv preprint arXiv:2212.03533},
  year={2022}
}

@inproceedings{li2021active,
  title = {Active Feature Acquisition with Generative Surrogate Models},
  author = {Li, Yang and Oliva, Junier},
  booktitle = {Proceedings of the 38th International Conference on Machine Learning},
  pages = {6450--6459},
  year = {2021}
}

@inproceedings{socher2013recursive,
  title = {Recursive Deep Models for Semantic Compositionality Over a Sentiment Treebank},
  author = {Socher, Richard and Perelygin, Alex and Wu, Jean and Chuang, Jason and Manning, Christopher D. and Ng, Andrew and Potts, Christopher},
  booktitle = {Proceedings of the 2013 Conference on Empirical Methods in Natural Language Processing},
  pages = {1631--1642},
  year = {2013}
}

@inproceedings{zhang2015character,
  title = {Character-Level Convolutional Networks for Text Classification},
  author = {Zhang, Xiang and Zhao, Junbo and LeCun, Yann},
  booktitle = {Advances in Neural Information Processing Systems},
  year = {2015}
}

@inproceedings{saravia2018carer,
  title = {CARER: Contextualized Affect Representations for Emotion Recognition},
  author = {Saravia, Elvis and Liu, Hsien-Chi Toby and Huang, Yen-Hao and Wu, Junlin and Chen, Yi-Shin},
  booktitle = {Proceedings of the 2018 Conference on Empirical Methods in Natural Language Processing},
  pages = {3687--3697},
  year = {2018}
}

@inproceedings{li2010contextual,
  title = {A Contextual-Bandit Approach to Personalized News Article Recommendation},
  author = {Li, Lihong and Chu, Wei and Langford, John and Schapire, Robert E.},
  booktitle = {Proceedings of the 19th International Conference on World Wide Web},
  pages = {661--670},
  year = {2010}
}

@inproceedings{abbasi2011improved,
  title = {Improved Algorithms for Linear Stochastic Bandits},
  author = {Abbasi-Yadkori, Yasin and P{\'a}l, D{\'a}vid and Szepesv{\'a}ri, Csaba},
  booktitle = {Advances in Neural Information Processing Systems},
  volume = {24},
  year = {2011}
}

@inproceedings{agrawal2013thompson,
  title = {Thompson Sampling for Contextual Bandits with Linear Payoffs},
  author = {Agrawal, Shipra and Goyal, Navin},
  booktitle = {Proceedings of the 30th International Conference on Machine Learning},
  pages = {127--135},
  year = {2013}
}

@article{javanmard2019dynamic,
  title = {Dynamic Pricing in High-Dimensions},
  author = {Javanmard, Adel and Nazerzadeh, Hamid},
  journal = {Journal of Machine Learning Research},
  volume = {20},
  number = {9},
  pages = {1--49},
  year = {2019}
}

@inproceedings{niculescu2005predicting,
  title = {Predicting Good Probabilities with Supervised Learning},
  author = {Niculescu-Mizil, Alexandru and Caruana, Rich},
  booktitle = {Proceedings of the 22nd International Conference on Machine Learning},
  pages = {625--632},
  year = {2005}
}

@inproceedings{guo2017calibration,
  title = {On Calibration of Modern Neural Networks},
  author = {Guo, Chuan and Pleiss, Geoff and Sun, Yu and Weinberger, Kilian Q.},
  booktitle = {Proceedings of the 34th International Conference on Machine Learning},
  pages = {1321--1330},
  year = {2017}
}

@inproceedings{kazerouni2017conservative,
  title = {Conservative Contextual Linear Bandits},
  author = {Kazerouni, Abbas and Ghavamzadeh, Mohammad and Abbasi-Yadkori, Yasin and Van Roy, Benjamin},
  booktitle = {Proceedings of the 20th International Conference on Artificial Intelligence and Statistics},
  pages = {1481--1490},
  year = {2017}
}

@inproceedings{sui2015safe,
  title = {Safe Exploration for Optimization with Gaussian Processes},
  author = {Sui, Yanan and Gotovos, Alkis and Burdick, Joel and Krause, Andreas},
  booktitle = {Proceedings of the 32nd International Conference on Machine Learning},
  pages = {997--1005},
  year = {2015}
}

@inproceedings{amani2019linear,
  title = {Linear Stochastic Bandits Under Safety Constraints},
  author = {Amani, Sanae and Alizadeh, Mahnoosh and Thrampoulidis, Christos},
  booktitle = {Advances in Neural Information Processing Systems},
  volume = {32},
  year = {2019}
}

@article{cohn1996active,
  title = {Active Learning with Statistical Models},
  author = {Cohn, David A. and Ghahramani, Zoubin and Jordan, Michael I.},
  journal = {Journal of Artificial Intelligence Research},
  volume = {4},
  pages = {129--145},
  year = {1996}
}

@techreport{settles2009active,
  title = {Active Learning Literature Survey},
  author = {Settles, Burr},
  institution = {University of Wisconsin--Madison},
  number = {1648},
  year = {2009}
}

@inproceedings{koh2017understanding,
  title = {Understanding Black-Box Predictions via Influence Functions},
  author = {Koh, Pang Wei and Liang, Percy},
  booktitle = {Proceedings of the 34th International Conference on Machine Learning},
  pages = {1885--1894},
  year = {2017}
}

@inproceedings{pruthi2020estimating,
  title = {Estimating Training Data Influence by Tracing Gradient Descent},
  author = {Pruthi, Garima and Liu, Frederick and Kale, Satyen and Sundararajan, Mukund},
  booktitle = {Advances in Neural Information Processing Systems},
  volume = {33},
  pages = {19920--19930},
  year = {2020}
}

@inproceedings{dwork2006calibrating,
  title = {Calibrating Noise to Sensitivity in Private Data Analysis},
  author = {Dwork, Cynthia and McSherry, Frank and Nissim, Kobbi and Smith, Adam},
  booktitle = {Proceedings of the Third Theory of Cryptography Conference},
  pages = {265--284},
  year = {2006}
}

@article{pedregosa2011scikit,
  title = {Scikit-learn: Machine Learning in Python},
  author = {Pedregosa, Fabian and Varoquaux, Ga{\"e}l and Gramfort, Alexandre and Michel, Vincent and Thirion, Bertrand and Grisel, Olivier and Blondel, Mathieu and Prettenhofer, Peter and Weiss, Ron and Dubourg, Vincent and others},
  journal = {Journal of Machine Learning Research},
  volume = {12},
  pages = {2825--2830},
  year = {2011}
}

\section{Formal Setup, Algorithms, and Baselines}
\label{app:formal_method}
\label{app:formal}

This appendix gives the formal interaction protocol, the full \textsc{NH-CROP} decision rule, and the information-access assumptions for all baselines. 
The main text presents the core ideas; here we make the implementation-level details explicit.

\subsection{Notation and Interaction Protocol}
\label{app:notation_protocol}

Table~\ref{tab:notation} summarizes the main notation.

\begin{table*}[t]
\centering
\small
\begin{tabular}{lp{0.72\linewidth}}
\toprule
Symbol & Meaning \\
\midrule
$x_t$ & NLP task context at round $t$. \\
$d_t$ & Candidate language data asset. \\
$\tilde c_t$ & Coarse privacy/access-cost estimate observed before pricing. \\
$c_t^\star$ & True privacy/access cost, hidden from non-oracle policies before pricing. \\
$(\mu_t,\sigma_t)$ & Cost-belief mean and uncertainty. \\
$v_t$ & Verification decision, where $v_t=1$ means paying for a refined cost signal. \\
$c_{\mathrm{ver}}$ & Verification cost. \\
$s_t$ & Refined cost-related signal obtained after verification. \\
$p_t \in \mathcal{P}$ & Posted price chosen from a discrete price set. \\
$y_t$ & Binary purchase feedback. \\
$r_t$ & Realized safe net reward. \\
$\hat q_t(p,c)$ & Estimated purchase probability at price $p$ using cost proxy $c$. \\
$b_t(p,c)$ & Contextual optimism bonus. \\
$\bar q_t(p,c)$ & Clipped optimistic purchase-probability estimate. \\
$\widehat R_t(p,c)$ & Estimated clipped safe-revenue score. \\
$q_{\max}$ & Shared clipping value selected on validation seeds. \\
$\lambda$ & Risk parameter for conservative cost proxy $\mu_t+\lambda\sigma_t$. \\
$\gamma$ & No-harm margin for triggering verification. \\
$V_t^{\mathrm{dir}}, V_t^{\mathrm{risk}}, V_t^{\mathrm{ver}}$ & Estimated values of direct pricing, risk-aware pricing, and verify-then-price. \\
$\Delta_t^{\mathrm{info}}$ & Counterfactual value of exact cost information, used only for diagnostics. \\
\bottomrule
\end{tabular}
\caption{
Notation used in the paper.
}
\label{tab:notation}
\end{table*}

At each round, the platform observes $(x_t,d_t,\tilde c_t)$, optionally verifies the asset, posts a price $p_t$, observes purchase feedback $y_t$, and receives
\begin{equation}
    r_t = y_t(p_t-c_t^\star)-c_{\mathrm{ver}}v_t .
    \label{eq:app_reward}
\end{equation}
The objective is cumulative safe net revenue, $\sum_{t=1}^T r_t$.

\subsection{Cost Belief and Demand Model}
\label{app:belief_demand}

The coarse and refined cost signals are modeled as
\begin{equation}
    \tilde c_t = c_t^\star+\epsilon_t,
    \qquad
    s_t = c_t^\star+\eta_t ,
    \label{eq:app_cost_signals}
\end{equation}
where $\eta_t$ is typically lower-variance than $\epsilon_t$.
Let
\begin{equation}
    z_t =
    \begin{cases}
    s_t, & v_t=1,\\
    \tilde c_t, & v_t=0 .
    \end{cases}
\end{equation}
The cost-belief mean is updated by
\begin{equation}
    \mu_{t+1}=(1-\alpha)\mu_t+\alpha z_t ,
    \label{eq:app_ema}
\end{equation}
and the uncertainty update is
\begin{equation}
    \sigma_{t+1}
    =
    \begin{cases}
    \sigma_{\mathrm{ver}}, & v_t=1,\\
    \max(\rho\sigma_t,\sigma_{\mathrm{unver}}), & v_t=0 .
    \end{cases}
    \label{eq:app_sigma_update}
\end{equation}
The floor $\sigma_{\mathrm{unver}}$ prevents coarse metadata from unrealistically eliminating uncertainty.

For a candidate price $p$ and cost proxy $c$, the contextual demand model uses features
\begin{equation}
    \phi_t(p,c)=\phi(x_t,d_t,p,c)
\end{equation}
and estimates purchase probability as
\begin{equation}
    \hat q_t(p,c)
    =
    \sigma_{\mathrm{logit}}
    \left(
    \hat\theta_t^\top \phi_t(p,c)
    \right).
\end{equation}
We use a standard contextual optimism bonus
\begin{equation}
    b_t(p,c)
    =
    \beta_t
    \sqrt{
    \phi_t(p,c)^\top V_t^{-1}\phi_t(p,c)
    } .
    \label{eq:app_bonus}
\end{equation}
After observing $y_t$, the design matrix is updated as
\begin{equation}
    V_{t+1}=V_t+\psi_t\psi_t^\top,
\end{equation}
where $\psi_t=\phi_t(p_t,\hat c_t)$ and $\hat c_t$ is the cost proxy used for pricing. 
The demand parameter is updated by online regularized logistic regression.

\subsection{\textsc{NH-CROP} Decision Rule}
\label{app:nhcrop_rule}

\paragraph{Clipped robust pricing.}
To avoid over-aggressive optimism under cost uncertainty, \textsc{NH-CROP} clips the optimistic purchase estimate:
\begin{equation}
    \bar q_t(p,c)
    =
    \mathrm{clip}
    \left(
    \hat q_t(p,c)+b_t(p,c),0,q_{\max}
    \right).
    \label{eq:app_clipped_q}
\end{equation}
The estimated safe-revenue score is
\begin{equation}
    \widehat R_t(p,c)=\bar q_t(p,c)(p-c).
    \label{eq:app_safe_score}
\end{equation}
The same validation-selected $q_{\max}$ is used for all clipped methods, including clipped Price-Only and Risk-Averse baselines.

\paragraph{No-verification alternatives.}
The direct action prices from the current cost belief:
\begin{equation}
    V_t^{\mathrm{dir}}
    =
    \max_{p\in\mathcal{P}}
    \widehat R_t(p,\mu_t).
\end{equation}
The risk-aware action prices from a conservative cost proxy:
\begin{equation}
    c_t^{\mathrm{risk}}=\mu_t+\lambda\sigma_t,
    \qquad
    V_t^{\mathrm{risk}}
    =
    \max_{p\in\mathcal{P}}
    \widehat R_t(p,c_t^{\mathrm{risk}}).
\end{equation}

\paragraph{Verify-then-price value.}
Let $\mathcal{S}_t$ denote the predictive distribution over refined cost signals. 
Using $K$ Monte Carlo samples $\tilde s_t^{(k)}\sim\mathcal{S}_t$, we estimate
\begin{equation}
    V_t^{\mathrm{ver}}
    =
    \frac{1}{K}
    \sum_{k=1}^{K}
    \max_{p\in\mathcal{P}}
    \widehat R_t(p,\tilde s_t^{(k)})
    -
    c_{\mathrm{ver}} .
    \label{eq:app_verify_value}
\end{equation}
The no-harm gate verifies only if
\begin{equation}
    v_t
    =
    \mathbf{1}
    \left[
    V_t^{\mathrm{ver}}
    >
    \max(V_t^{\mathrm{dir}},V_t^{\mathrm{risk}})+\gamma
    \right].
    \label{eq:app_no_harm_gate}
\end{equation}
If the gate rejects verification, the platform chooses the better of direct and risk-aware pricing.
If the gate accepts verification, it observes $s_t$, updates the cost belief, and prices from the refined estimate.
Thus, zero verification can be the intended behavior when refined information has low estimated decision value.

\subsection{Algorithm}
\label{app:algorithm}

\refstepcounter{algorithm}
\begin{center}
\fbox{
\begin{minipage}{0.95\textwidth}
\small
\textbf{Algorithm~\thealgorithm: \textsc{NH-CROP}}\\[0.25em]
\textbf{Input:} price set $\mathcal{P}$, verification cost $c_{\mathrm{ver}}$, clipping value $q_{\max}$, risk parameter $\lambda$, no-harm margin $\gamma$, Monte Carlo count $K$.\\
\textbf{Initialize:} demand model $\hat\theta_1$, design matrix $V_1$, asset-level cost beliefs $(\mu_d,\sigma_d)$.

\begin{enumerate}
    \item[] \textbf{For} $t=1,\ldots,T$:
    \begin{enumerate}
        \item Observe $(x_t,d_t,\tilde c_t)$ and retrieve cost belief $(\mu_t,\sigma_t)$.
        \item Compute $V_t^{\mathrm{dir}}$ from $\mu_t$ and $V_t^{\mathrm{risk}}$ from $\mu_t+\lambda\sigma_t$.
        \item Estimate $V_t^{\mathrm{ver}}$ using sampled refined signals and subtract $c_{\mathrm{ver}}$.
        \item If $V_t^{\mathrm{ver}}>\max(V_t^{\mathrm{dir}},V_t^{\mathrm{risk}})+\gamma$, verify, observe $s_t$, update the cost belief, and price from the refined estimate.
        \item Otherwise, skip verification and post the price from the better no-verification action.
        \item Observe $y_t$, receive $r_t$, and update the demand model and cost belief.
    \end{enumerate}
\end{enumerate}
\end{minipage}
}
\label{alg:nhcrop}
\end{center}

\FloatBarrier

\subsection{Decision-Relevance Diagnostic}
\label{app:decision_relevance_definition}

For post-hoc analysis, we define the counterfactual value of exact cost information:
\begin{equation}
    \Delta_t^{\mathrm{info}}
    =
    \max_{p\in\mathcal{P}}\widehat R_t(p,c_t^\star)
    -
    \max_{p\in\mathcal{P}}\widehat R_t(p,\mu_t).
    \label{eq:app_info_value}
\end{equation}
This quantity is not available to non-oracle policies because it depends on $c_t^\star$.
It is used only to analyze whether cost information would have changed the pricing decision.
A large estimation error $|\mu_t-c_t^\star|$ does not necessarily imply large $\Delta_t^{\mathrm{info}}$; verification matters only when the refined information changes a consequential decision.

\subsection{Baselines and Information Access}
\label{app:baseline_access}

Table~\ref{tab:baseline_access} summarizes what each method can access.
Oracle policies are diagnostic upper bounds and are not deployable.

\begin{table*}[t]
\centering
\small
\begin{tabular}{lccccc}
\toprule
Method & Coarse cost & Clipping & Risk fallback & Can verify & Oracle/hindsight \\
\midrule
Price-Only UCB & Yes & No & No & No & No \\
Price-Only Clipped UCB & Yes & Yes & No & No & No \\
Risk-Averse UCB & Yes & No & Yes & No & No \\
Risk-Averse Clipped UCB & Yes & Yes & Yes & No & No \\
TPIV-UCB & Yes & No & No & Yes & No \\
\textsc{NH-CROP} Full & Yes & Optional & Yes & Yes & No \\
\textsc{NH-CROP} Full-Clipped & Yes & Yes & Yes & Yes & No \\
\textsc{NH-CROP} Clip-NoV & Yes & Yes & Yes & No & No \\
Always Verify & Yes & No & No & Yes & No \\
Random Verify & Yes & No & No & Yes & No \\
Free Oracle & Yes & Optional & Optional & Yes & Yes \\
Oracle Positive ROI & Yes & Optional & Optional & Yes & Yes \\
Oracle Price-Change Positive & Yes & Optional & Optional & Yes & Yes \\
\bottomrule
\end{tabular}
\caption{
Information access of non-oracle and oracle methods.
Non-oracle methods never observe $c_t^\star$ before pricing.
Oracle policies use hindsight information and are included only as upper bounds.
}
\label{tab:baseline_access}
\end{table*}

\paragraph{Purpose of key baselines.}
Price-Only UCB tests whether pricing from coarse costs is sufficient.
Risk-Averse UCB tests whether conservative no-verification pricing can replace information acquisition.
TPIV-UCB tests the simpler rule of verifying whenever $\sigma_t>\tau$.
\textsc{NH-CROP} Clip-NoV disables verification while keeping the clipped robust pricing structure, isolating whether gains come from calibration or from actual paid verification.
Oracle baselines measure the potential value of cost information when useful verification events can be selected with hindsight.
\section{Benchmark Construction and Reproducibility}
\label{app:benchmarks_reproducibility}
\label{app:benchmarks}

This appendix summarizes the benchmark construction, validation protocol, and reproducibility setup.
The synthetic benchmark is a controlled stress test, while the real-proxy and utility-grounded benchmarks use real language-data slices with proxy costs or measured downstream utility.
None of the proxy costs should be interpreted as legal, contractual, or deployed marketplace costs.

\subsection{Benchmark Overview}
\label{app:benchmark_overview}

Table~\ref{tab:benchmark_overview} summarizes the five evaluation settings.
All main results use 30 seeds.

\begin{table*}[t]
\centering
\small
\begin{tabular}{llrp{0.55\linewidth}}
\toprule
Setting & Source & Rounds & Purpose \\
\midrule
\textsc{SYN-high}
& Synthetic market
& 420
& Controlled high-cost-uncertainty setting for testing robust pricing and optional information acquisition. \\

\textsc{RP-base}
& Real text slices with proxy costs
& 260
& Default real-proxy setting using text-derived and metadata-derived privacy/access-cost proxies. \\

\textsc{RP-high-DV}
& Real text slices with high decision value
& 260
& Stress test where access-cost information is more likely to affect pricing decisions. \\

\textsc{UT-base}
& Downstream utility matrix
& 260
& Pricing setting where asset value is tied to measured downstream NLP utility. \\

\textsc{UT-high}
& Utility matrix with high tradeoff pressure
& 260
& Stress test emphasizing high-utility/high-cost and low-utility/low-cost tradeoffs. \\
\bottomrule
\end{tabular}
\caption{
Benchmark settings.
The real-proxy and utility-grounded settings use real language-data slices, but the privacy/access costs are proxy variables rather than marketplace-observed costs.
}
\label{tab:benchmark_overview}
\end{table*}

\subsection{Synthetic Market}
\label{app:synthetic_market}

The synthetic market isolates the interaction between task-conditioned demand, uncertain access costs, and optional verification.
Each round samples a task context $x_t$, an asset $d_t$, a latent cost $c_t^\star$, and a coarse estimate
\begin{equation}
    \tilde c_t = c_t^\star + \epsilon_t .
\end{equation}
The high-uncertainty setting uses the largest coarse-estimation noise level.
To avoid trivial uncertainty collapse under repeated observations, the latent cost may drift slowly:
\begin{equation}
    c_t^\star = c_{t-1}^\star + \xi_t .
\end{equation}
If verification is performed, the platform receives a lower-variance signal
\begin{equation}
    s_t = c_t^\star + \eta_t .
\end{equation}

Task contexts include task type, budget level, and privacy sensitivity.
Assets include source/domain, quality, size, rarity, and task-affinity features.
Purchase feedback is sampled from a logistic demand model:
\begin{equation}
\Pr(y_t=1)
=
\sigma_{\mathrm{logit}}
\left(
\beta_0
+
\beta_{\mathrm{rel}}\mathrm{rel}(x_t,d_t)
+
\beta_q q(d_t)
-
\beta_p \rho(x_t)p_t
-
\beta_c \kappa(x_t)\hat c_t
\right),
\label{eq:app_synthetic_purchase}
\end{equation}
where $\mathrm{rel}(x_t,d_t)$ is task--asset relevance, $q(d_t)$ is asset quality, $\rho(x_t)$ controls price sensitivity, $\kappa(x_t)$ controls cost sensitivity, and $\hat c_t$ is the platform's current cost proxy.

\subsection{Real-Proxy Benchmark}
\label{app:real_proxy}

The real-proxy benchmark constructs candidate assets from SST-2, AG News, and an emotion classification dataset \citep{socher2013recursive,zhang2015character,saravia2018carer}.
Each dataset is partitioned into small language-data slices.
Each slice is associated with source identity, sample count, label distribution, text-length statistics, class imbalance, label entropy, and quality indicators.

The true privacy/access-cost proxy is computed from normalized feature groups:
\begin{equation}
    c^\star(d)
    =
    \mathrm{clip}
    \left(
    \sum_{j=1}^{m} w_j g_j(d),
    0,
    1
    \right),
    \label{eq:app_real_proxy_cost}
\end{equation}
where $g_j(d)$ are text-derived or metadata-derived proxy features.
Table~\ref{tab:real_proxy_components} reports the grouped cost features used in the paper-facing benchmark.

\begin{table*}[t]
\centering
\small
\begin{tabular}{lp{0.56\linewidth}r}
\toprule
Proxy group & Examples of included signals & Weight \\
\midrule
Sensitive-pattern risk
& Email-like strings, phone-like strings, URL/IP-like strings, numeric identifiers
& 0.30 \\

Duplication / contamination risk
& Exact duplicate ratio, repeated $n$-gram ratio, approximate near-duplicate score
& 0.25 \\

Toxicity or sensitive-content proxy
& Lexicon-based sensitive indicators and source-level risk proxy
& 0.15 \\

License / access prior
& Source-level access prior and dataset-level reuse assumptions
& 0.15 \\

Quality risk
& Empty-text rate, malformed-text rate, length outliers, class imbalance
& 0.10 \\

Rarity / size risk
& Small-slice rarity and domain specificity
& 0.05 \\
\bottomrule
\end{tabular}
\caption{
Proxy groups used to construct real-proxy privacy/access costs.
These proxies are designed for controlled pricing experiments and are not privacy or legal guarantees.
}
\label{tab:real_proxy_components}
\end{table*}

The platform does not observe the full proxy cost at pricing time.
It receives a coarse estimate from cheap metadata such as source identity, slice size, average length, and label entropy:
\begin{equation}
    \tilde c(d) = h_{\mathrm{coarse}}(d) + \epsilon .
\end{equation}
Verification simulates a lightweight audit that reveals a refined estimate from a subset of text-derived proxy features:
\begin{equation}
    s(d) = h_{\mathrm{verify}}(d) + \eta .
\end{equation}

\subsection{Downstream-Utility-Grounded Benchmark}
\label{app:utility_grounded}

The utility-grounded benchmark ties asset value to downstream NLP performance.
For each task $a$ and candidate asset $d$, we compute utility as validation improvement from adding the asset to a fixed base training set:
\begin{equation}
    u(a,d)
    =
    \mathrm{Score}
    \left(
    \mathrm{base}(a) \cup d
    \right)
    -
    \mathrm{Score}
    \left(
    \mathrm{base}(a)
    \right).
    \label{eq:app_utility}
\end{equation}
We use TF--IDF features and logistic regression implemented with scikit-learn \citep{pedregosa2011scikit}.
This lightweight setup keeps the benchmark reproducible and CPU-friendly.
The task families are sentiment, topic, and emotion classification, evaluated with accuracy or macro-F1 depending on the task.

The simulator generates purchase feedback from utility, price, and cost:
\begin{equation}
\Pr(y_t=1)
=
\sigma_{\mathrm{logit}}
\left(
\beta_0
+
\beta_u u(x_t,d_t)
-
\beta_p \rho(x_t)p_t
-
\beta_c \kappa(x_t)\hat c_t
+
\epsilon_t
\right).
\label{eq:app_utility_demand}
\end{equation}
The \textsc{UT-high} variant emphasizes high-utility/high-cost and low-utility/low-cost tradeoffs.

\subsection{High Decision-Value Variants}
\label{app:high_decision_value}

The base real-proxy and utility-grounded settings can have low practical value of verification: even refined cost information may not change the final pricing decision enough to justify information acquisition.
We therefore include high decision-value variants.
\textsc{RP-high-DV} increases the influence of access-cost information on demand and safe revenue, especially for assets whose coarse estimate lies near the pricing margin.
\textsc{UT-high} emphasizes utility--cost tradeoffs where high-value assets may also carry high proxy cost.
These variants are defined by fixed environment parameters before policy evaluation, rather than selected post-hoc based on method performance.

\subsection{Hyperparameters and Validation Protocol}
\label{app:hyperparameters}

Hyperparameters that affect policy selection are chosen on validation seeds and fixed for evaluation.
Table~\ref{tab:hyperparams} reports the main settings.

\begin{table*}[t]
\centering
\small
\begin{tabular}{llp{0.52\linewidth}}
\toprule
Parameter & Value & Notes \\
\midrule
Evaluation seeds
& 30
& Used for all main paired comparisons. \\

Synthetic horizon
& 420 rounds
& Used for \textsc{SYN-high}. \\

Real-proxy horizon
& 260 rounds
& Used for \textsc{RP-base} and \textsc{RP-high-DV}. \\

Utility-grounded horizon
& 260 rounds
& Used for \textsc{UT-base} and \textsc{UT-high}. \\

Price grid
& $\{0.1,0.2,\ldots,1.0\}$
& Shared by all policies. \\

Clipping value $q_{\max}$
& 0.8
& Validation-selected and shared by all clipped methods. \\

Monte Carlo samples $K$
& configured value
& Used to estimate verify-then-price value. \\

Risk parameter $\lambda$
& validation-selected
& Used for risk-aware cost proxy $\mu_t+\lambda\sigma_t$. \\

No-harm margin $\gamma$
& validation-selected
& Used by the information-acquisition gate. \\

Verification cost $c_{\mathrm{ver}}$
& setting-specific
& Fixed within each environment. \\

Seed-level comparison
& directional paired test
& Computed over seed-level cumulative safe net revenue. \\
\bottomrule
\end{tabular}
\caption{
Main hyperparameters and evaluation protocol.
}
\label{tab:hyperparams}
\end{table*}

The clipping value is tuned once and shared by clipped Price-Only UCB, Risk-Averse UCB, and \textsc{NH-CROP}.
Table~\ref{tab:clip_tuning} reports the validation summary.

\begin{table}[t]
\centering
\small
\begin{tabular}{rr}
\toprule
Clip value $q_{\max}$ & Validation mean reward \\
\midrule
0.8 & 0.0622 \\
0.5 & 0.0570 \\
1.2 & 0.0557 \\
0.3 & 0.0525 \\
0.2 & 0.0512 \\
0.1 & 0.0512 \\
\bottomrule
\end{tabular}
\caption{
Clip-value validation summary.
}
\label{tab:clip_tuning}
\end{table}

\subsection{Statistical Reporting and Reproducibility}
\label{app:reproducibility}

For each setting and method, we report seed count, number of rounds, cumulative safe net revenue, mean reward per round, verification frequency, verification ROI statistics when applicable, and price-change-after-verification rate when applicable. Primary comparisons are paired by seed. Directional $p$-values are used only for pre-specified comparisons, while appendix diagnostics report additional robustness summaries where available.

The implementation is organized into modules for environments, asset construction, agents, evaluation, and experiment scripts.
The paper-facing outputs are generated from saved CSV and trajectory files:
\begin{itemize}
    \item \texttt{tables/final\_setting\_method\_summary.csv};
    \item \texttt{tables/method\_independent\_relevance\_buckets.csv};
    \item \texttt{tables/final\_method\_independent\_stratified.csv};
    \item \texttt{tables/clip\_tuning\_summary.csv};
    \item \texttt{raw/seed\_level\_results.csv};
    \item \texttt{raw/round\_level\_results.csv}.
\end{itemize}
The final paper-ready audit is reproduced with:
\begin{verbatim}
python -m src.experiments.run_emnlp_final_audit --full
\end{verbatim}
Earlier diagnostic runs use the corresponding experiment-audit and verification-contribution-audit scripts in the released repository.

The benchmarks are intentionally lightweight.
The synthetic and real-proxy settings run without large-scale GPU training, and the utility-grounded benchmark uses TF--IDF logistic regression.
The proxy costs are intended to create controlled and interpretable uncertainty for pricing experiments; they do not replace privacy auditing, legal review, or deployed marketplace pricing data.
\section{Full Empirical Results and Robustness Analyses}
\label{app:full_results}
\label{app:supplementary-results}

This appendix expands the empirical results in Section~\ref{sec:experiments}.
We focus on four diagnostics: full non-oracle results, fair clipped baseline comparisons, method-independent decision-relevance buckets, and controlled synthetic sweeps.
Information-acquisition audits and oracle verification analyses are reported separately in Appendix~\ref{app:verification_audit}.

\subsection{Full Non-Oracle Results}
\label{app:full_main_results}

Table~\ref{tab:app_full_main_results} reports cumulative safe net revenue for the main non-oracle methods.
The clipped \textsc{NH-CROP} variants improve over Price-Only UCB in all five settings and are strongest or competitive among learned non-oracle methods.
In \textsc{UT-base}, Risk-Averse UCB is slightly stronger than \textsc{NH+Clip}, so we interpret the result as evidence for robust calibration rather than universal dominance.

\begin{table}[!htbp]
\centering
\small
\setlength{\tabcolsep}{4pt}
\begin{tabular}{lrrrrrr}
\toprule
Setting
& Price
& Risk
& NH
& NH+Clip
& NH+Clip-NoV
& $p_{\mathrm{NH+Clip}}$ vs Price \\
\midrule
SYN-high
& 20.05
& 20.00
& 23.88
& 25.45
& \textbf{25.68}
& $<.001$ \\

RP-base
& 35.63
& 36.00
& 37.59
& \textbf{38.01}
& \textbf{38.01}
& .001 \\

RP-high-DV
& 20.59
& 20.87
& 22.13
& \textbf{23.42}
& \textbf{23.42}
& $<.001$ \\

UT-base
& 4.96
& \textbf{5.45}
& 5.09
& 5.40
& 5.40
& .018 \\

UT-high
& 5.08
& 5.02
& 5.13
& \textbf{5.41}
& \textbf{5.41}
& .112 \\
\bottomrule
\end{tabular}
\caption{
Full non-oracle result summary.
Values are cumulative safe net revenue averaged over 30 seeds.
``Price'' denotes Price-Only UCB, ``Risk'' denotes Risk-Averse UCB, ``NH'' denotes \textsc{NH-CROP}, and ``NH+Clip-NoV'' disables verification while retaining clipped robust pricing.
The final column reports directional paired $p$-values for the pre-specified comparison of \textsc{NH+Clip} against Price-Only UCB.
}
\label{tab:app_full_main_results}
\end{table}

\subsection{Fair Clipped Baseline Comparison}
\label{app:fair_clipped}

A potential concern is that \textsc{NH-CROP} benefits only because it receives an additional clipping parameter.
To address this, the same validation-selected clipping value, $q_{\max}=0.8$, is applied to Price-Only UCB, Risk-Averse UCB, and \textsc{NH-CROP}.
Table~\ref{tab:app_fair_clipping} reports the effect of applying clipping within each method family.

Clipping is not a universal improvement.
It weakens Price-Only and Risk-Averse baselines in most settings, while improving \textsc{NH-CROP} variants.
This suggests that clipping is most useful when combined with the no-harm pricing structure rather than as a generic post-hoc adjustment.

\begin{table}[!htbp]
\centering
\small
\begin{tabular}{lrrrr}
\toprule
Setting
& Price+Clip $-$ Price
& Risk+Clip $-$ Risk
& NH+Clip $-$ NH
& NH+Clip-NoV $-$ NH-NoV \\
\midrule
SYN-high
& -0.67
& -1.31
& +1.57
& +1.87 \\

RP-base
& -1.19
& -1.54
& +0.42
& +0.42 \\

RP-high-DV
& -0.85
& -1.07
& +1.29
& +1.29 \\

UT-base
& -0.02
& -0.51
& +0.31
& +0.31 \\

UT-high
& -0.47
& -0.24
& +0.27
& +0.27 \\
\bottomrule
\end{tabular}
\caption{
Effect of applying the shared clipping protocol.
Values are differences in cumulative safe net revenue.
Clipping often weakens Price-Only and Risk-Averse baselines but improves the \textsc{NH-CROP} family.
}
\label{tab:app_fair_clipping}
\end{table}

\subsection{Method-Independent Decision-Relevance Buckets}
\label{app:decision_buckets}
\label{app:decision-relevance}

We stratify rounds by a method-independent decision-relevance score.
The score is computed from the Price-Only UCB trajectory using the counterfactual value of replacing the platform's current cost belief with the true cost.
Rounds are divided into low, medium, and high buckets by quantiles of this score, and all methods are evaluated on the same bucketed rounds.

\begin{figure}[!htbp]
\centering
\includegraphics[width=0.96\linewidth]{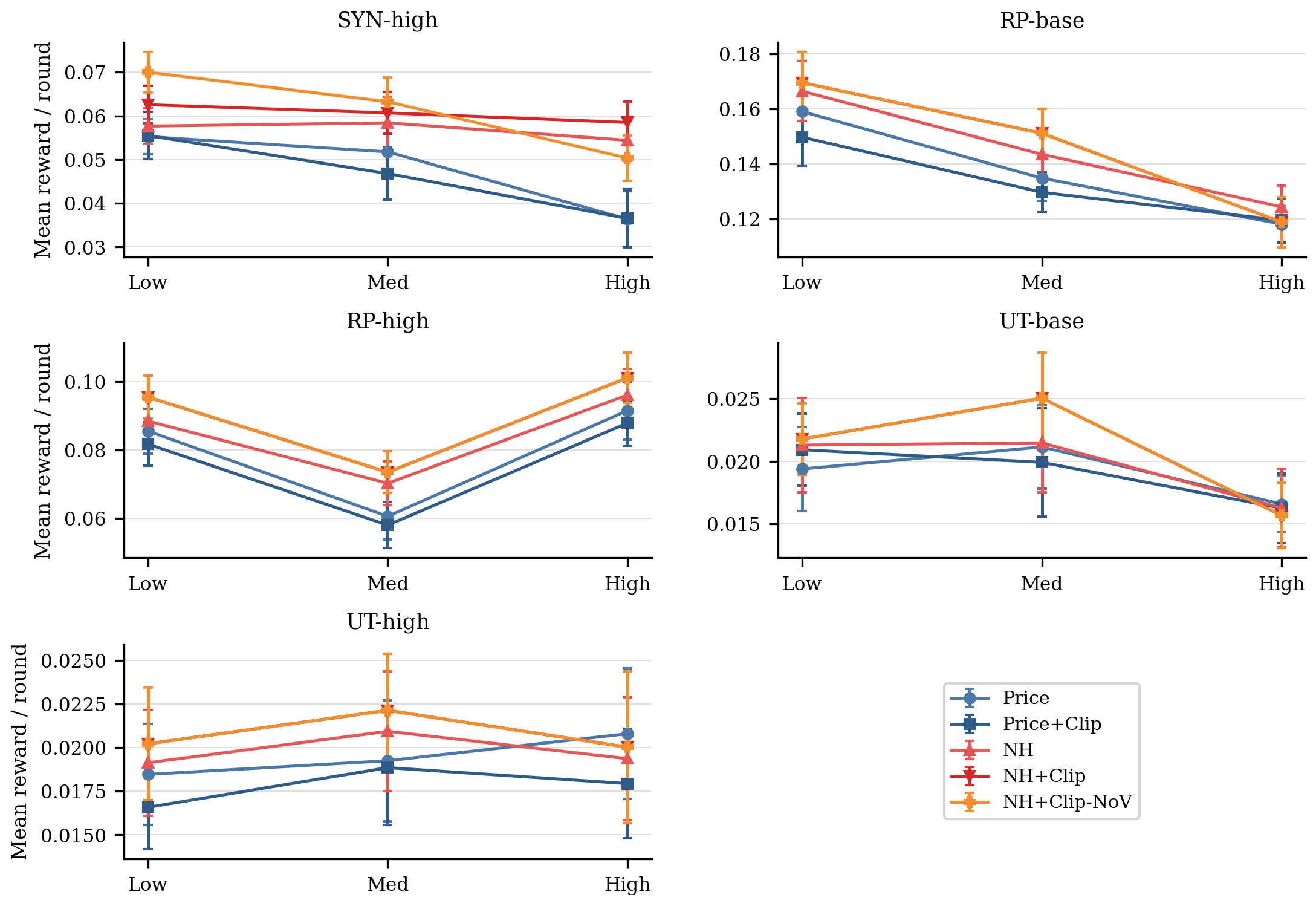}
\caption{
Full method-independent decision-relevance stratification across all settings.
Each panel reports mean safe net reward per round within low-, medium-, and high-relevance buckets.
The pattern is clearest in real-proxy and synthetic settings and weaker in utility-grounded settings.
}
\label{fig:app_decision_buckets}
\end{figure}

Table~\ref{tab:app_decision_buckets} gives a compact version of the bucketed results.
The pattern is strongest in real-proxy and synthetic settings: clipped \textsc{NH-CROP} variants often improve over Price-Only UCB in low- and medium-relevance buckets, while the utility-grounded settings are weaker and less consistent.
Figure~\ref{fig:app_decision_buckets} visualizes the same comparison.

\begin{table}[!htbp]
\centering
\small
\setlength{\tabcolsep}{4pt}
\begin{tabular}{llrrrr}
\toprule
Setting & Bucket & Price & NH & NH+Clip & NH+Clip-NoV \\
\midrule
SYN-high & Low
& 0.0553 & 0.0577 & 0.0626 & \textbf{0.0700} \\
SYN-high & Medium
& 0.0518 & 0.0584 & 0.0607 & \textbf{0.0632} \\
SYN-high & High
& 0.0364 & 0.0544 & \textbf{0.0585} & 0.0503 \\

\midrule
RP-base & Low
& 0.1589 & 0.1663 & \textbf{0.1694} & \textbf{0.1694} \\
RP-base & Medium
& 0.1347 & 0.1434 & \textbf{0.1510} & \textbf{0.1510} \\
RP-base & High
& 0.1180 & \textbf{0.1243} & 0.1187 & 0.1187 \\

\midrule
RP-high-DV & Low
& 0.0855 & 0.0885 & \textbf{0.0955} & \textbf{0.0955} \\
RP-high-DV & Medium
& 0.0606 & 0.0703 & \textbf{0.0735} & \textbf{0.0735} \\
RP-high-DV & High
& 0.0916 & 0.0961 & \textbf{0.1011} & \textbf{0.1011} \\

\midrule
UT-base & Low
& 0.0194 & 0.0213 & \textbf{0.0218} & \textbf{0.0218} \\
UT-base & Medium
& 0.0211 & 0.0215 & \textbf{0.0250} & \textbf{0.0250} \\
UT-base & High
& \textbf{0.0166} & 0.0163 & 0.0157 & 0.0157 \\

\midrule
UT-high & Low
& 0.0185 & 0.0191 & \textbf{0.0202} & \textbf{0.0202} \\
UT-high & Medium
& 0.0192 & 0.0209 & \textbf{0.0221} & \textbf{0.0221} \\
UT-high & High
& \textbf{0.0208} & 0.0194 & 0.0200 & 0.0200 \\
\bottomrule
\end{tabular}
\caption{
Method-independent decision-relevance stratification.
Entries are mean reward per round within each bucket.
Buckets are defined from a shared Price-Only UCB trajectory, so all methods are evaluated on the same low-, medium-, and high-relevance round sets.
}
\label{tab:app_decision_buckets}
\end{table}

\subsection{Component Ablations}
\label{app:ablations}

Table~\ref{tab:app_component_ablations} isolates two components: clipping and the no-verification fallback.
Disabling verification usually does not hurt performance because the full method rarely verifies in real-proxy and utility-grounded settings.
By contrast, adding clipping consistently improves the \textsc{NH-CROP} family.
This supports the main interpretation that robust pricing calibration is more important than actual paid verification in the current benchmarks.

\begin{table}[!htbp]
\centering
\small
\begin{tabular}{lrrrrrr}
\toprule
Setting
& Full
& NoV
& Full+Clip
& Clip-NoV
& Full$-$NoV
& Clip-NoV$-$NoV \\
\midrule
SYN-high
& 23.88 & 23.82 & 25.45 & 25.68 & +0.07 & +1.87 \\

RP-base
& 37.59 & 37.59 & 38.01 & 38.01 & 0.00 & +0.42 \\

RP-high-DV
& 22.13 & 22.13 & 23.42 & 23.42 & 0.00 & +1.29 \\

UT-base
& 5.09 & 5.09 & 5.40 & 5.40 & 0.00 & +0.31 \\

UT-high
& 5.13 & 5.13 & 5.41 & 5.41 & 0.00 & +0.27 \\
\bottomrule
\end{tabular}
\caption{
Component ablation for no-verification fallback and clipping.
Values are cumulative safe net revenue.
The full and no-verification variants are nearly identical in real-proxy and utility-grounded settings, while clipping improves the \textsc{NH-CROP} family across all settings.
}
\label{tab:app_component_ablations}
\end{table}

\subsection{Controlled Synthetic Robustness Sweeps}
\label{app:controlled_sweeps}

We additionally report controlled robustness sweeps from the earlier uncertainty-triggered TPIV-UCB simulator.
These sweeps use a different reward scale from the final \textsc{NH-CROP} audit and should be interpreted within-figure only.
They are included as mechanism diagnostics rather than as the primary empirical claim of the paper.

\begin{figure}[!htbp]
\centering
\includegraphics[width=\linewidth]{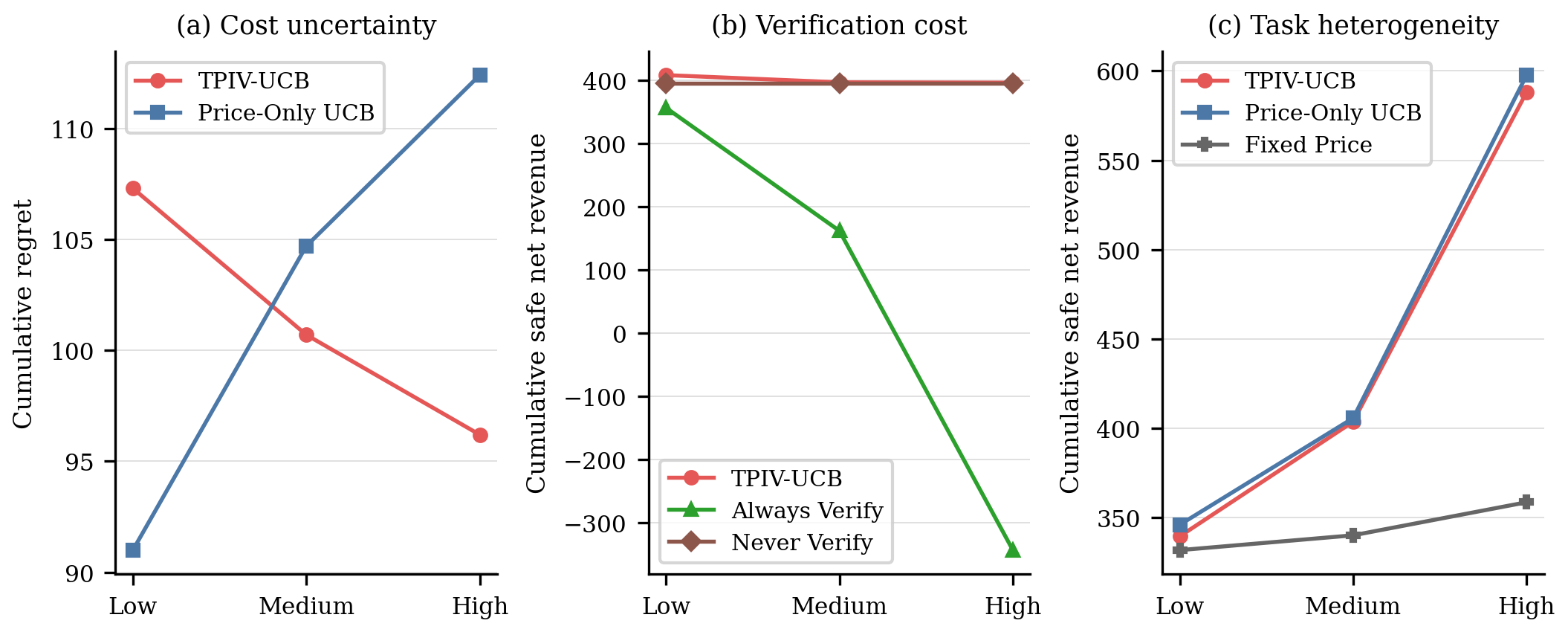}
\caption{
Controlled synthetic robustness sweeps for the uncertainty-triggered TPIV-UCB baseline.
Panel A varies cost-estimation uncertainty, Panel B varies verification cost, and Panel C varies task heterogeneity.
These sweeps are mechanism diagnostics and should be interpreted separately from the final \textsc{NH-CROP} audit.
}
\label{fig:app_synthetic_sweeps}
\end{figure}

Figure~\ref{fig:app_synthetic_sweeps} summarizes three sweeps.
Panel A varies coarse cost-estimation uncertainty and shows a synthetic crossover: uncertainty-triggered verification becomes more useful only when uncertainty is high.
Panel B varies verification cost and shows that Always Verify collapses as verification becomes expensive.
Panel C varies task heterogeneity and shows that contextual pricing matters, but uncertainty-triggered verification does not consistently outperform Price-Only UCB in that sweep.

\subsection{Additional Visual Summaries}
\label{app:additional_visual_summaries}

\begin{figure}[!htbp]
\centering
\includegraphics[width=0.96\linewidth]{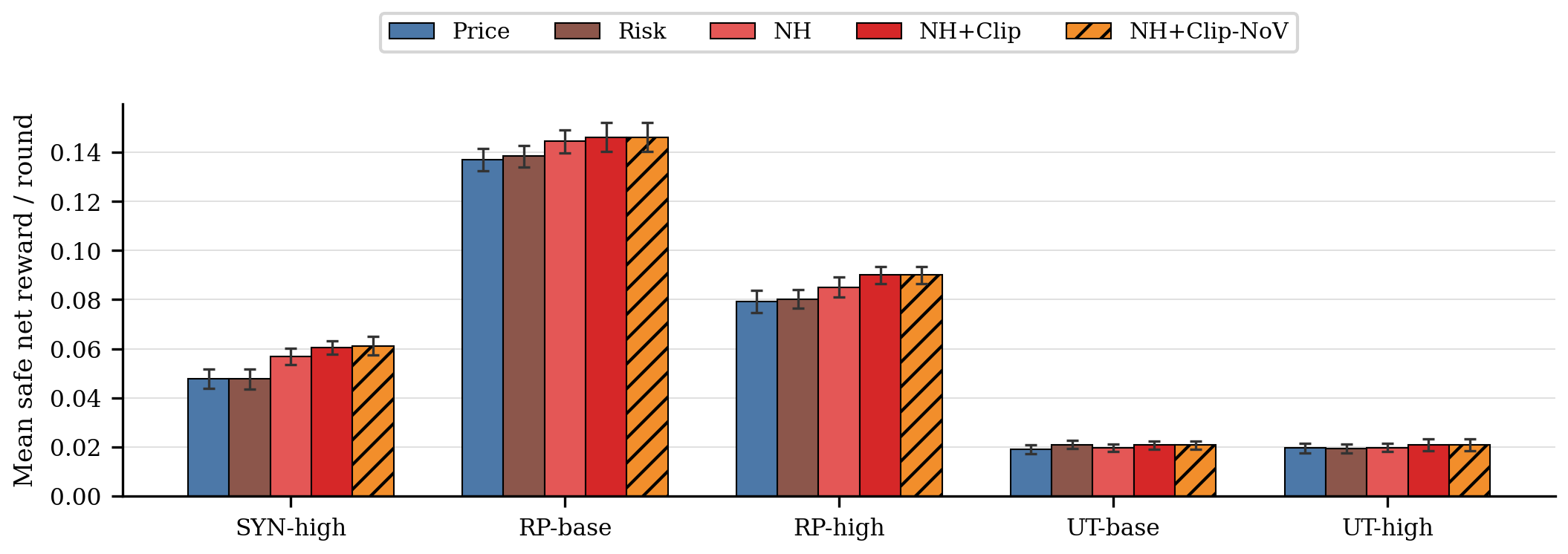}
\caption{
Main selected fair-clipped pricing results.
Bars show mean safe net reward per round averaged over 30 seeds; error bars denote 95\% confidence intervals.
The figure visualizes the selected learned-policy comparison, while Table~\ref{tab:main_results} reports the full fair-clipped baseline results including Price+Clip and Risk+Clip.
The hatched NH+Clip-NoV bars disable verification, illustrating that robust pricing calibration is the dominant practical driver.
}
\label{fig:app_main_results}
\end{figure}

\begin{figure}[!htbp]
\centering
\includegraphics[width=0.96\linewidth]{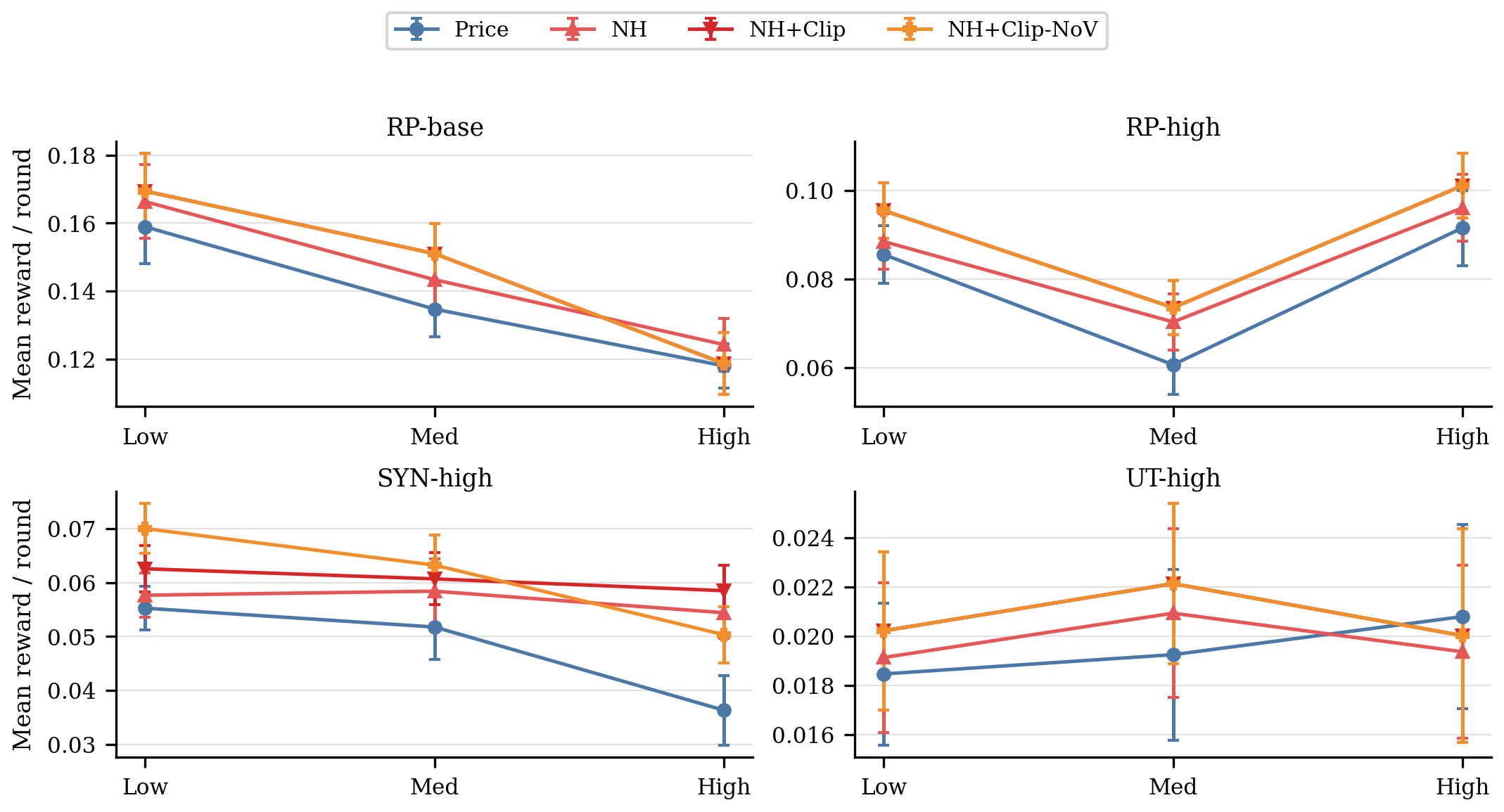}
\caption{
Compact method-independent decision-relevance stratification.
Rounds are bucketed using a shared Price-Only UCB trajectory and counterfactual information value, so all methods are evaluated on the same low-, medium-, and high-relevance rounds.
This compact view complements the full stratification in Figure~\ref{fig:app_decision_buckets}.
}
\label{fig:app_decision_relevance_compact}
\end{figure}

These additional visual summaries are included at the end of Appendix~C so that the full stratification remains Figure~\ref{fig:app_decision_buckets} and the compact view remains an appendix-only complement.

\FloatBarrier

\subsection{Summary of Appendix C}
\label{app:full_results_summary}

The supplementary results support three conclusions.
First, clipped \textsc{NH-CROP} variants improve over Price-Only UCB across all main settings and are strongest or competitive among learned non-oracle methods.
Second, fair clipping does not simply help every method; it helps the \textsc{NH-CROP} family more than Price-Only or Risk-Averse baselines.
Third, decision-relevance stratification and controlled synthetic sweeps support the diagnostic claim that cost information is useful only when it changes pricing decisions, while practical gains in the main benchmarks should not be attributed to actual paid verification without a causal audit.
\section{Information-Acquisition Audit, Oracle Bounds, and Case Studies}
\label{app:verification_audit}
\label{app:verification-audit}

This appendix audits the role of information acquisition.
The main result is that actual paid verification is not the dominant source of learned-policy gains in the real-proxy or utility-grounded settings.
However, oracle baselines show that refined cost information can still have substantial potential value.
The gap between oracle value and learned verification value motivates the no-harm interpretation of \textsc{NH-CROP}: verification should be optional and decision-value-dependent.

\subsection{Causal Verification Ablation}
\label{app:causal_verification}

Table~\ref{tab:d1_causal_verification_ablation} compares the full policy with no-verification and no-cost-verification variants.
If paid verification drove the gains, the full policy should outperform the no-verification variant.
This is not what we observe.
In real-proxy and utility-grounded settings, the full policy either matches its no-verification counterpart or differs negligibly, with zero verification frequency.
The only nonzero verification frequency appears in \textsc{SYN-high}, where the contribution is small.
This supports the interpretation that robust clipped pricing, not actual paid verification, is the main practical driver.

\begin{table}[!htbp]
\centering
\small
\setlength{\tabcolsep}{4pt}
\begin{tabular}{lrrrrrr}
\toprule
Setting
& Full
& NoV
& Full+Clip
& Clip-NoV
& No-Cost Verif.
& $v$-Freq. \\
\midrule
SYN-high
& 23.88
& 23.82
& 25.45
& \textbf{25.68}
& 24.98
& 0.026 \\

RP-base
& 37.59
& 37.59
& \textbf{38.01}
& \textbf{38.01}
& 36.81
& 0.000 \\

RP-high-DV
& 22.13
& 22.13
& \textbf{23.42}
& \textbf{23.42}
& 21.76
& 0.000 \\

UT-base
& 5.09
& 5.09
& \textbf{5.40}
& \textbf{5.40}
& 5.23
& 0.000 \\

UT-high
& 5.13
& 5.13
& \textbf{5.41}
& \textbf{5.41}
& 5.15
& 0.000 \\
\bottomrule
\end{tabular}
\caption{
Causal verification ablation.
Values are cumulative safe net revenue averaged over 30 seeds.
``NoV'' disables verification while preserving the pricing structure, and ``Clip-NoV'' disables verification while retaining clipped robust pricing.
``No-Cost Verif.'' removes verification cost from the learned verification behavior; it is distinct from oracle policies, which use hindsight information.
The full policy does not outperform the no-verification variant in real-proxy or utility-grounded settings.
}
\label{tab:d1_causal_verification_ablation}
\end{table}

Table~\ref{tab:d1_causal_verification_ablation} summarizes the causal verification audit; we omit the redundant small-multiple plot for readability.

\subsection{Verification ROI and VOI Calibration}
\label{app:voi_calibration}

Table~\ref{tab:app_verification_event_stats} summarizes event-level verification behavior.
No-cost verification reveals that useful local events exist, especially in synthetic and real-proxy settings.
However, the estimated-VOI trigger over-verifies: it often changes prices but yields negative realized ROI.
This explains why verification can have oracle value while learned verification policies fail to exploit it reliably.
Figure~\ref{fig:oracle-information-value} (D.2) visualizes the resulting gap between oracle information value and learned gains.

\begin{table}[!htbp]
\centering
\small
\setlength{\tabcolsep}{4pt}
\begin{tabular}{llrrrr}
\toprule
Setting
& Policy
& Changed price
& Positive ROI
& Mean ROI
& Interpretation \\
\midrule
RP-base
& No-cost verification
& 0.669
& 0.669
& 0.0245
& useful local events \\

RP-base
& Est-VOI trigger
& 0.540
& 0.098
& -0.0341
& over-verifies \\

RP-high-DV
& No-cost verification
& 0.605
& 0.605
& 0.0136
& useful local events \\

RP-high-DV
& Est-VOI trigger
& 0.569
& 0.079
& -0.0386
& over-verifies \\

SYN-high
& Full policy
& 0.805
& 0.140
& -0.0283
& rare and noisy \\

SYN-high
& No-cost verification
& 0.577
& 0.577
& 0.0063
& weak local value \\

SYN-high
& Est-VOI trigger
& 0.514
& 0.010
& -0.0446
& over-verifies \\

UT-base
& No-cost verification
& 0.566
& 0.566
& 0.0045
& small local value \\

UT-base
& Est-VOI trigger
& 0.469
& 0.000
& -0.0467
& no positive ROI \\

UT-high
& No-cost verification
& 0.491
& 0.491
& 0.0042
& small local value \\

UT-high
& Est-VOI trigger
& 0.401
& 0.000
& -0.0471
& no positive ROI \\
\bottomrule
\end{tabular}
\caption{
Verification event statistics.
``Changed price'' is the fraction of verified rounds in which the refined signal changes the posted price.
``Positive ROI'' is the fraction of verified rounds with positive realized verification ROI.
No-cost verification identifies useful local events, while the estimated-VOI trigger is miscalibrated and often over-verifies.
}
\label{tab:app_verification_event_stats}
\end{table}

\setcounter{figure}{1}

\begin{figure}[!htbp]
\centering
\includegraphics[width=\linewidth]{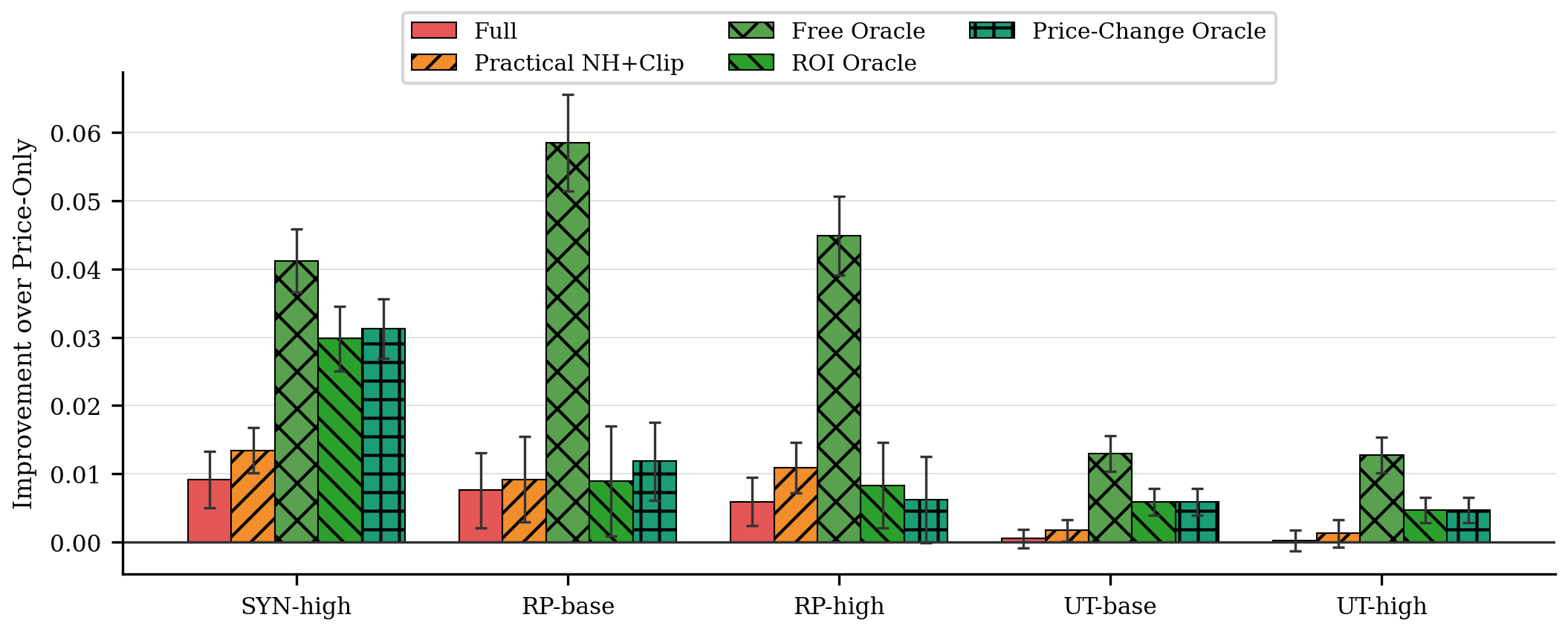}
\caption{
Oracle information value versus learned gains.
Oracle policies are diagnostic upper bounds and are not deployable. They reveal substantial potential value in refined cost information, but learned policies do not reliably identify useful verification events before paying for them.
}
\label{fig:oracle-information-value}
\end{figure}

\subsection{Representative Verification Cases}
\label{app:case_studies}

\begin{figure}[!htbp]
\centering
\includegraphics[width=\linewidth]{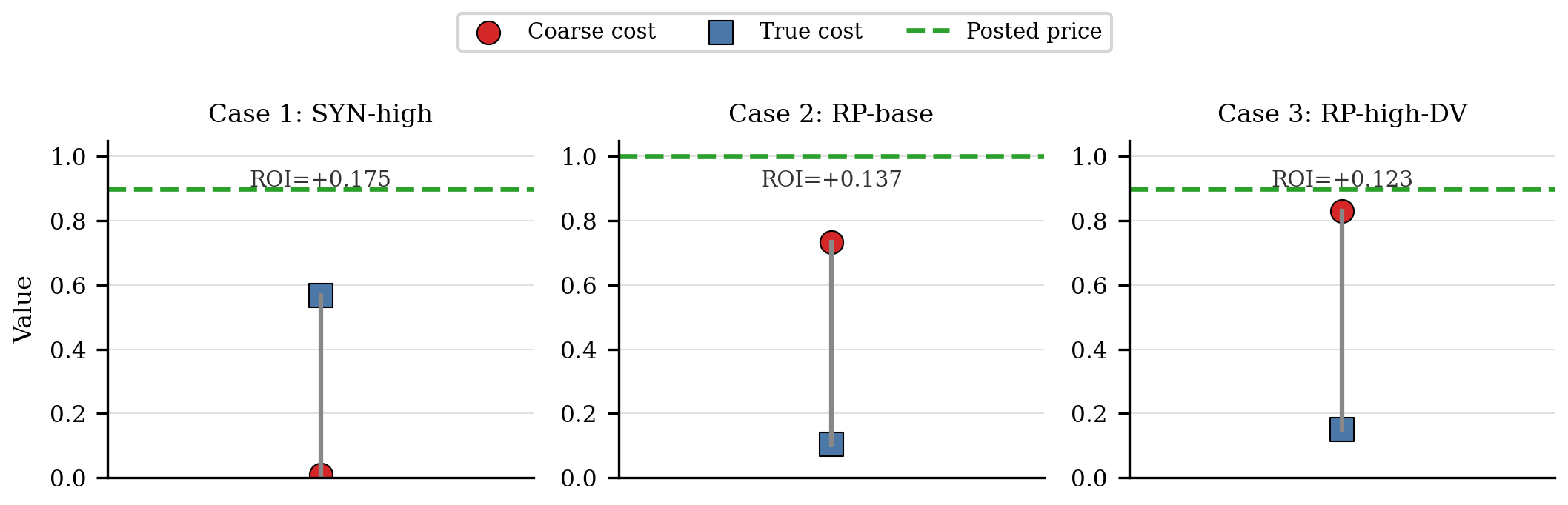}
\caption{
Representative positive verification events.
Each panel compares the coarse cost estimate, true cost, and posted price for a selected round with positive realized verification ROI.
}
\label{fig:representative-verification-cases}
\end{figure}

Table~\ref{tab:d3_representative_cases} and Figure~\ref{fig:representative-verification-cases} show representative positive verification events.
These cases are selected from verified rounds in which refined information changes the price and yields positive realized ROI.
They illustrate that useful verification opportunities exist, but they should not be interpreted as evidence that the learned full policy obtains its aggregate gains from verification.

\begin{table}[!htbp]
\centering
\small
\setlength{\tabcolsep}{4pt}
\begin{tabular}{llrrrrrrr}
\toprule
Setting
& Method
& Seed
& Round
& Price
& True cost
& Cost est.
& Est. VOI
& ROI \\
\midrule
SYN-high
& No-cost verification
& 26
& 54
& 0.90
& 0.492
& 0.000
& 0.219
& 0.142 \\

SYN-high
& No-cost verification
& 10
& 10
& 0.20
& 0.518
& 0.000
& 0.233
& 0.139 \\

RP-base
& No-cost verification
& 6
& 177
& 1.00
& 0.075
& 0.945
& 0.004
& 0.139 \\

RP-base
& No-cost verification
& 17
& 11
& 1.00
& 0.101
& 0.734
& 0.049
& 0.138 \\

RP-base
& No-cost verification
& 5
& 232
& 1.00
& 0.090
& 0.871
& 0.008
& 0.137 \\

RP-high-DV
& No-cost verification
& 15
& 74
& 0.80
& 0.152
& 0.768
& 0.037
& 0.124 \\
\bottomrule
\end{tabular}
\caption{
Representative positive verification events.
These examples show that refined cost information can correct large coarse-cost errors and improve local realized reward.
They are illustrative events, not aggregate evidence that paid verification drives the main learned-policy gains.
}
\label{tab:d3_representative_cases}
\end{table}

\subsection{Failure Modes}
\label{app:failure_modes}

The audit identifies four recurring failure modes.
First, verification can change the selected price without improving realized reward, so price-change rate alone is not sufficient evidence of value.
Second, estimated VOI can be miscalibrated, causing aggressive verification policies to over-verify and incur negative ROI.
Third, oracle information value does not imply learned verification value, because deployable policies must identify useful events before paying for them.
Fourth, no verification can be the correct action in low-information-value regimes.
These failure modes explain why \textsc{NH-CROP} treats verification as optional and no-harm rather than as the default response to uncertainty.

\subsection{Summary}
\label{app:verification_audit_summary}

The information-acquisition audit supports three conclusions.
First, actual paid verification does not drive the main gains in real-proxy or utility-grounded settings.
Second, clipped robust pricing is the dominant practical mechanism among learned policies.
Third, oracle analyses show that refined cost information can have large potential value, leaving open the challenge of learning better value-of-information policies.
\section{Additional Robustness Checks}
\label{app:additional-robustness}

This appendix reports two additional robustness checks designed to stress-test the main interpretation. They are diagnostic rather than new primary benchmarks. We do not change the main environment parameters, and all thresholds or calibrators are selected using validation seeds only. Evaluation seeds are used only for reporting. Negative results are retained.

\subsection{Transformer Utility Sanity Check}
\label{app:transformer-utility}

The original utility-grounded benchmark uses TF--IDF logistic regression so that the full audit remains reproducible and CPU-friendly. To test whether this makes the utility conclusions too dependent on a lightweight model, we reconstruct the utility matrix using transformer representations from \texttt{intfloat/e5-small-v2} (\citealp{wang2022text}). The experiment uses three task families and 720 candidate assets, with four validation seeds and eight evaluation seeds. All reported runs used the transformer backend.

Table~\ref{tab:transformer-utility-diagnostics} summarizes the utility-matrix diagnostics. The transformer utility distribution is weakly correlated with the original lightweight utility distribution: Pearson correlation is 0.0316 and Spearman correlation is 0.0678. The fraction of positive-utility assets also changes substantially, from 0.6708 under the original utility matrix to 0.2083 under the transformer-derived utility matrix. Thus, this sanity check is not merely a repeated report of the same utility distribution.

Despite this distribution shift, the pricing conclusion remains unchanged. In both UT-TRANS-base and UT-TRANS-high, NH+Clip and NH+Clip-NoV coincide and perform zero verification. CalVOI does not improve over the no-verification fallback: its gap versus NoV is -1.0262 in UT-TRANS-base and -0.4435 in UT-TRANS-high. This supports the main interpretation that robust no-verification pricing remains the safer learned behavior in low-actionability utility-grounded regimes.

\begin{table}[!htbp]
\centering
\small
\begin{tabular}{l r}
\toprule
Diagnostic & Value \\
\midrule
Number of tasks & 3 \\
Number of assets & 720 \\
Original utility mean / std & 0.01187 / 0.02066 \\
Transformer utility mean / std & 0.00102 / 0.01460 \\
Original vs Transformer Pearson & 0.0316 \\
Original vs Transformer Spearman & 0.0678 \\
Positive utility fraction, Transformer & 0.2083 \\
Positive utility fraction, original & 0.6708 \\
Utility--cost correlation, Transformer & 0.0445 \\
Utility--cost correlation, original & 0.1520 \\
\bottomrule
\end{tabular}
\caption{Transformer-utility sanity-check diagnostics. The transformer-derived utility matrix differs substantially from the original lightweight utility distribution, but the learned pricing conclusion remains unchanged.}
\label{tab:transformer-utility-diagnostics}
\end{table}

\begin{table}[!htbp]
\centering
\small
\begin{tabular}{l r r r r}
\toprule
Setting & NH+Clip-NoV & NH+Clip v-freq. & CalVOI gap vs NoV & $p$ \\
\midrule
UT-TRANS-base & 0.025065 & 0.000 & -1.0262 & 0.0673 \\
UT-TRANS-high & 0.020601 & 0.000 & -0.4435 & 0.3876 \\
\bottomrule
\end{tabular}
\caption{Transformer-utility pricing sanity check. Entries for NH+Clip-NoV are mean reward per round. CalVOI gaps are cumulative-reward differences versus NH+Clip-NoV. Paid verification does not improve over the no-verification fallback in these utility-grounded settings.}
\label{tab:transformer-utility-pricing}
\end{table}

\subsection{CalVOI Feature Ablation and Generalization}
\label{app:calvoi-ablation}

CalVOI is evaluated only as a robustness diagnostic. It is a calibrated gate trained on validation seeds to select verification events from pre-verification features; it is not proposed as a replacement for the main NH-CROP policy. We next test whether the weakness of learned verification is merely due to overfitting a single CalVOI threshold. We evaluate CalVOI-full, feature ablations that remove price-gap, uncertainty, or task/source features, a VOI-only variant, and a cross-setting-threshold variant. All thresholds are selected on validation seeds. The evaluation settings are RP-base, GOV-HIGHVOI with verification costs 0.0050 and 0.0200, UT-EMB-base, and UT-EMB-high. UT-EMB-base and UT-EMB-high are supplementary embedding-utility settings used only for robustness diagnostics.

The results show a narrow positive window. In the high-VOI, low-verification-cost setting, CalVOI variants can outperform the no-verification fallback. The strongest ablation, CalVOI without uncertainty features, improves over NoV by +4.0449 cumulative reward with $p=0.0075$ and win rate 0.8. CalVOI-full improves by +2.3198 but is weaker statistically ($p=0.0760$). The cross-setting-threshold variant improves by +3.6866 with $p=0.0899$. These results indicate that useful learned verification is possible when refined cost information is cheap and decision-actionable.

However, this behavior does not generalize across regimes. At higher verification cost, CalVOI-full has a negative gap (-0.9469, $p=0.6081$), and the best ablation is only weakly positive and non-significant (+0.4934, $p=0.8166$). In RP-base, NH+Clip-NoV remains strongest; CalVOI-full has a negative gap (-3.6828, $p=0.0864$). In UT-EMB-base, CalVOI-full is significantly worse than NoV (-2.4846, $p=0.0070$), and in UT-EMB-high it is again non-improving. EVSI/estimated-VOI triggers are strongly negative in multiple settings, reinforcing that uncalibrated VOI estimates can over-verify.

\begin{table}[!htbp]
\centering
\small
\begin{tabular}{l l r r r}
\toprule
Setting & Method / comparison & Gap vs NoV & $p$ & Win rate \\
\midrule
GOV-HIGHVOI, $c_{\mathrm{ver}}=0.0050$ & CalVOI-no-uncertainty & +4.0449 & 0.0075 & 0.80 \\
GOV-HIGHVOI, $c_{\mathrm{ver}}=0.0050$ & CalVOI-full & +2.3198 & 0.0760 & 0.70 \\
GOV-HIGHVOI, $c_{\mathrm{ver}}=0.0050$ & CalVOI-cross-threshold & +3.6866 & 0.0899 & 0.60 \\
GOV-HIGHVOI, $c_{\mathrm{ver}}=0.0050$ & EVSI / estimated VOI & -3.9790 & 0.0143 & -- \\
GOV-HIGHVOI, $c_{\mathrm{ver}}=0.0200$ & CalVOI-full & -0.9469 & 0.6081 & -- \\
GOV-HIGHVOI, $c_{\mathrm{ver}}=0.0200$ & Best ablation & +0.4934 & 0.8166 & -- \\
RP-base & CalVOI-full & -3.6828 & 0.0864 & 0.40 \\
RP-base & EVSI / estimated VOI & -14.9646 & $9.05{\times}10^{-5}$ & -- \\
UT-EMB-base & CalVOI-full & -2.4846 & 0.0070 & -- \\
UT-EMB-high & CalVOI-full & -0.3436 & 0.3389 & -- \\
\bottomrule
\end{tabular}
\caption{CalVOI feature ablation and threshold-generalization summary. Gaps are cumulative-reward differences relative to NH+Clip-NoV. These ablations are diagnostic and are not proposed as separate deployable methods. CalVOI has a positive window only in the high-VOI, low-verification-cost setting; in RP-base, the supplementary embedding-utility settings, and higher-cost regimes, no-verification robust pricing remains stronger or competitive.}
\label{tab:calvoi-ablation}
\end{table}

\subsection{Takeaway from Additional Robustness Checks}

The appendix checks support two conservative conclusions. First, the main no-verification finding is not only a consequence of the original TF--IDF utility proxy: it persists under a transformer-derived utility matrix with substantially different utility statistics. Second, calibrated verification is not useless, but it is conditional. It can help in high-VOI, low-verification-cost regimes, yet it is unstable or harmful in RP-base, the supplementary embedding-utility settings, and high-cost settings. These results strengthen the paper's main claim that governed language-data platforms should calibrate robust pricing first and pay for additional information only when its decision value is actionable.

\end{document}